\documentclass[runningheads]{llncs}
\usepackage{graphicx}
\usepackage{tikz}
\usepackage{comment}
\usepackage{amsmath,amssymb} % define this before the line numbering.
\usepackage{color}

\usepackage{graphicx}
\usepackage{epsfig}
\usepackage{amssymb}
\usepackage{multirow}
\usepackage{subfigure}
\usepackage{emptypage}
\usepackage{verbatimbox}
\usepackage{booktabs}

\usepackage{color}
\usepackage{footmisc}
\usepackage[misc]{ifsym}
\usepackage{cite}
\usepackage{color}
\usepackage[colorlinks,linkcolor=red]{hyperref}
\usepackage[accsupp]{axessibility}  % Improves PDF readability for those with disabilities.

\begin{document}
% \linenumbers
\pagestyle{headings}
\mainmatter
\def\ECCVSubNumber{6460}  % Insert your submission number here

\title{CODER: Coupled Diversity-Sensitive Momentum Contrastive Learning \\ for Image-Text Retrieval} % Replace with your title

\begin{comment}
\titlerunning{ECCV-20 submission ID \ECCVSubNumber} 
\authorrunning{ECCV-20 submission ID \ECCVSubNumber} 
\author{Anonymous ECCV submission}
\institute{Paper ID \ECCVSubNumber}
\end{comment}
%******************

%\begin{comment}
\titlerunning{CODER}
\author{Haoran Wang\inst{1} \and
Dongliang He\inst{1}\thanks{indicates corresponding author.} \and
Wenhao Wu\inst{1,5} \and
Boyang Xia\inst{2} \and
Min Yang\inst{1} \and
Fu Li\inst{1} \and
Yunlong Yu\inst{3} \and
Zhong Ji\inst{4} \and
Errui Ding\inst{1} \and
Jingdong Wang\inst{1}
}
\authorrunning{Haoran Wang et al.}
\institute{Department of Computer Vision Technology (VIS), Baidu Inc., Beijing, China\\ 
\and
Key Lab of Intelligent Information Processing of Chinese Academy of Sciences (CAS), Institute of Computing Technology, CAS, Beijing, China\\
%\email{xiaboyang20@mails.ucas.ac.cn}\\
\and
College of Information Science \& Electronic Engineering, Zhejiang University, Hangzhou, China\\
%\email{yuyunlong@zju.edu.cn}\\
\and
School of Electrical \& Information Engineering, Tianjin University, Tianjin, China\\
%\email{jizhong@tju.edu.cn}\\
\and
The University of Sydney, Sydney, Australia\\
\email{\{wanghaoran09,hedongliang01,yangmin09,lifu\}@baidu.com,\\
xiaboyang20@mails.ucas.ac.cn,yuyunlong@zju.edu.cn,jizhong@tju.edu.cn,\\
% \{dingerrui,wangjingdong\}@baidu.com
}
}

%\end{comment}
%******************
\maketitle

%%%%%%%%% ABSTRACT
\begin{abstract}		
	Image-Text Retrieval (ITR) is challenging in bridging visual and lingual modalities. Contrastive learning has been adopted by most prior arts. Except for limited amount of negative image-text pairs, the capability of constrastive learning is restricted by manually weighting negative pairs as well as unawareness of external knowledge. In this paper, we propose our novel Coupled Diversity-Sensitive Momentum Constrastive Learning (CODER) for improving cross-modal representation. Firstly, a novel diversity-sensitive contrastive learning (DCL) architecture is invented. We introduce dynamic dictionaries for both modalities to enlarge the scale of image-text pairs, and diversity-sensitiveness is achieved by adaptive negative pair weighting. Furthermore, two branches are designed in CODER. One learns instance-level embeddings from image/text, and it also generates pseudo online clustering labels for its input image/text based on their embeddings. Meanwhile, the other branch learns to query from commonsense knowledge graph to form concept-level descriptors for both modalities. Afterwards, both branches leverage DCL to align the cross-modal embedding spaces while an extra pseudo clustering label prediction loss is utilized to promote concept-level representation learning for the second branch. Extensive experiments conducted on two popular benchmarks, \textit{i.e.} MSCOCO and Flicker30K, validate CODER remarkably outperforms the state-of-the-art approaches.	Our code is available at: \url{https://github.com/BruceW91/CODER}.		
\end{abstract}

%%%%%%%%% BODY TEXT
\section{Introduction}
\label{sec:intro}

Image-text retrieval (ITR) refers to searching for the semantically similar instance from visual (textual) modality with the query instance from textual (visual) modality. Nowadays, it has become a compelling topic from both industrial and research community and is of potential value to benefit extensive relevant applications	 \cite{2015VQA,xu2015show,ma2016learning,xu2017scene,bai2018deep,wang2018reconstruction,Gupta2019ViCoWE,hua2021exploiting,Zhao2021UnderstandingAE,jiao2021two,Yang2022UnifiedCL,Wu2022TransferringTK,jiao2022suspected}. In the past decade, tremendous progresses have been made with the prevalence of deep learning \cite{lecun2015deep}. Early works typically associate image with text via learning global \cite{2014UVSE,2016Wang,2018VSE++} or local cross-modal alignment \cite{2018SCAN,Chen2020IMRAMIM}. Follow-up studies attempt to introduce external knowledge information, including commonsense knowledge \cite{shi2019knowledge,Wang2020CVSE} or scene graph \cite{wang2020cross} information, into visual-semantic embedding models. It remains challenging due to heterogeneous multi-modal data distributions, which requires pretty precise cross-modal alignment.
	
\begin{figure}[t]		
	\begin{center}		
		\includegraphics[height=5.3cm,width=0.99\linewidth]{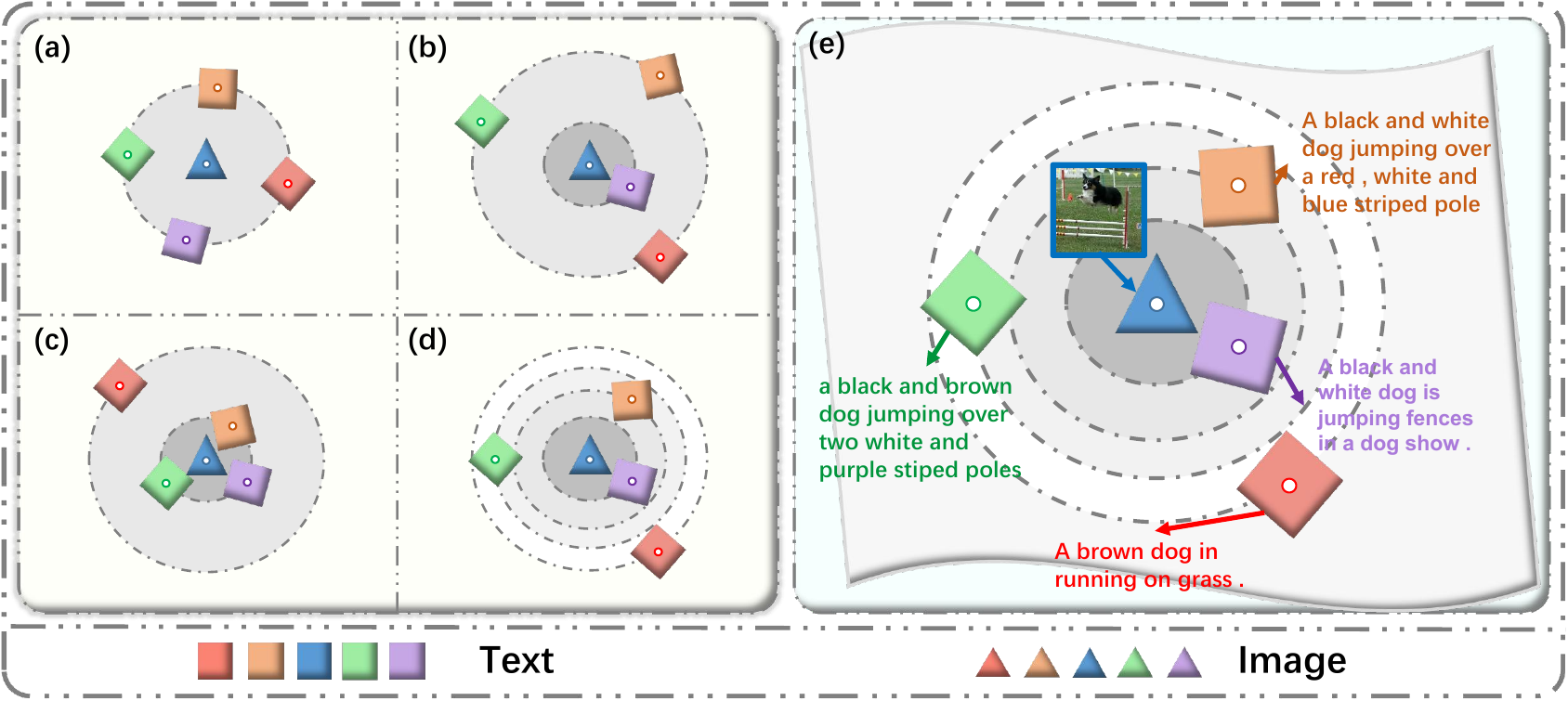}
	\end{center}	
	\caption{
		Conceptual illustration of our proposed Diversity-sensitive Contrastive Learning (DCL) loss. Sub-figure (a), (b) and (c) depict three exemplary distributions of negative samples which are undesired because they do not show much similarity variations, respectively. Sub-figure (d) shows desired negative sample distribution given an anchor, where different negative samples are not equally pushed away. It demonstrate the joint space can well distinguish fine-grained semantic difference among negative samples. Sub-figure (e) illustrates the ideal joint embedding space affected by DCL.  
	}		
	\label{fig.1}		
% 	\label{fig:long}			
% 	\label{fig:onecol}		
\end{figure}

Loss functions play the central role in aligning multi-modal data. The prevailing bi-directional triplet ranking (BTR) loss used in \cite{frome2013devise,2018VSE++} can be regarded as one special case of contrastive loss \cite{hadsell2006dimensionality}, where only one negative sample is considered. Then, bidirectional Info-NCE loss \cite{oord2018representation} (BIN), as a typical contrastive loss, has been widely adopted in many tasks \cite{radford2021learning,chen2020simple,Luo2021CLIP4ClipAE}. It exploits the whole paired relationships among a mini-batch of image-text samples when applied to the ITR task. Meanwhile, constrastive learning is well-known in limited negative sample scale \cite{he2020momentum}, which acts as the bottleneck of its capability. 

Another notable issue is both aforementioned contrastive losses manually design the weighting strategy for negative image-text pairs. They both enforce the negatives and anchor samples to be separated far away enough, whilst ignoring the relative differences between them. Consequently, the fine-grained discrepancies among negative pairs are hard to be fully captured.

In fact, the importance of each image/text instance is unequal \cite{chen2020adaptive} in contrastive learning. A critical factor determining the importance of instance is its semantic ambiguity \cite{2019PVSE}. In particular, the samples with high semantic ambiguity refers to those with multiple meanings/concepts. Oppositely, the samples with simple and clear meanings usually have low semantic ambiguity. To explicitly model the semantic ambiguity of sample, we present a term called ``\textbf{Diversity}''. Concretely, the diversity of one sample is defined based on the distribution of cross-modal negatives around it. For example, as depicted in three typical cases in Figure \ref{fig.1}(a-c), if a sample has multiple negatives with similar distances to it, we call this sample as low-diversity one. Obviously, the existence of low-diversity samples are undesirable, which will weaken the discrimination ability of the learned joint space. Conversely, if the data distribution around an anchor instance is well-spaced (see \ref{fig.1}(d), this sample has high diversity), it could better measure the difference among different negative samples, which is more ideal. 

To address the aforementioned limitations and questions, first of all, inspired by Momentum Contrastive Learning (MCL) paradigm \cite{he2020momentum}, dynamic dictionaries of memory banks are introduced in coupled form for both visual and textual modality to enlarge interactions among image-text pairs. Furthermore, in this paper, we propose to extend constrastive learning to a novel \textbf{D}iversity-sensitive \textbf{C}ontrastive \textbf{L}earning (\textbf{DCL}) paradigm. To achieve it, a novel diversity-sensitive contrastive loss is presented, which incorporates our defined diversity into contrastive loss. Specifically, in contrastive loss, a simple yet effective estimation function is designed to quantify the diversity of each anchor sample in a mini-batch of data, the diversity term is then used to dynamically weight negative samples of each anchor, enabling the training procedure to balance between diversity and total contrastive loss. With our DCL, on one hand, the image-text pairs built based on low diversity anchor sample can be allocated with larger weight and \textit{vice versa}; on the other hand, given a negative sample, when it is paired with different anchors, it can be unequally weighted according to the anchor's diversity. Doing so enables the original contrastive loss to be aware of semantic diversities of samples, and suppress the adverse impact brought by low-diversity ones. Accordingly, \textit{instance-level} visual or textual representations can be learned with our DCL. As consequence, we can obtain a more structured and hierarchical joint embedding space. Taking Figure \ref{fig.1}(e) as example, the subtle difference between the caption 
%``\texttt{A black and white dog jumping over a red, white and blue striped pole.}''
(marked in orange) and another one
%``\texttt{A black and brown dog jumping over two white and purple striped poles.}''
(marked in green) can be appropriately distinguished in their semantic distances.

Furthermore, how to leverage external knowledge into contrastive learning framework is worth exploring. To be complementary to the \textit{instance-level} alignment, we achieve \textit{concept-level} cross-modal feature alignment via exploiting commonsense knowledge. Different from the former, \textit{concept-level} alignment is built by firstly learning to extract homogeneous concept-level visual and textual embeddings from commonsense graph, followed by aligning the cross-modal embeddings via adopting DCL along with a \textbf{P}rototype-\textbf{G}uided \textbf{C}lassification loss (\textbf{PGC}). In order to enable PGC, an online clustering procedure is performed on \textit{instance-level}  representations and each cluster id is treated as a prototype, then a prediction head based on the \textit{concept-level} image/text embedding is employed for classifying the cluster id of the input image/text. The final image-text matching score is a combination of similarities obtained from both instance-level and concept-level alignment. Extensive experiments conducted on MSCOCO \cite{2014COCO} and Flicker30K \cite{2015Flickr30K} verify the superiority of our framework and show that our Coupled Diversity-Sensitive Contrastive Learning (CODER) method significantly outperforms recent state-of-the-art solutions.		

To sum up, the main contributions are listed as follows: 	

\begin{itemize}
	\item We incorporate coupled Momentum Contrastive learning (MCL) into image-text representation learning and further extend contrastive learning to a novel Diversity-Sensitive Contrastive Learning (DCL) paradigm, which can adaptively weight negative image-text pairs to further boost the performance. 
	
	\item A Coupled Diversity-Sensitive Contrastive Learning (CODER) framework is proposed to exploit not only instance-level image-text representations but also concept-level embeddings with the aid of external knowledge as well as on-line clustering based prototype-guided classification loss. 	
	
	\item Extensive experimental results on two benchmarks demonstrate our approach considerably outperforms state-of-the-art methods by a large margin.
\end{itemize}

\section{Related Work}
\label{sec:related_work}

\subsection{Contrastive Learning}
Recently, Contrastive Learning \cite{grill2020bootstrap,radford2021learning,oord2018representation,chen2020simple,he2020momentum} has made remarkable progress in unsupervised representation learning. Chen \textit{at el.} \cite{chen2020simple} shows that contrastive learning in unsupervised visual representation learning benefits from large batch size negatives and stronger data augmentation. He \textit{at el.} \cite{he2020momentum} proposed Momentum Contrastive Learning (MCL) paradigm that obtains the new key representation on-the-fly by a momentum-updated key encoder, and maintains a dictionary as a queue to allow the training process to reuse the encoded key representations from the immediate preceding mini-batches. Recently, more Contrastive Learning based vision-language understanding studies \cite{radford2021learning,huo2021wenlan,Li2021UNIMOTU,Zhang2021VLDeformerLV} are emerging. For video-text retrieval, Liu \textit{at el.} \cite{liu2021hit} first introduces the vanilla info-NCE loss based MCL mechanism to enhance the cross-modal discrimination. Distinct from them, we integrate coupled MCL into our proposed Diversity-sensitive contrastive learning (DCL) paradigm for tackling ITR.

\subsection{Image-Text Retrieval}
Along with the renaissance of deep learning, a surge of works have been proposed for ITR. Early attempts \cite{frome2013devise,2015mrnn,2015mcnn,2016Wang} typically employ global features to represent both image and text in a common semantic space. For instance, Kiros \textit{at el.} \cite{2014UVSE} encoded image and text by CNN and RNN respectively, utilizing BTR loss to train the model. Afterwards, another line of research \cite{2018SCAN,wehrmann2020adaptive,Wei2020MultiModalityCA,Chen2020IMRAMIM,diao2021similarity} employed multi-modal attention mechanism \cite{2018SCAN,Chen2020IMRAMIM,Wen2021LearningDS,ji2021step} or 
knowledge aided representation learning \cite{2018SCO,shi2019knowledge,liu2020graph,Wang2020CVSE,ge2021structured} to achieve cross-modal alignment by exploiting more fine-grained associations. For instance, Lee \emph{et al.} \cite{2018SCAN} developed Stacked Cross Attention Network that aligns image regions and textual words. 		

Except for focusing on representation architecture designing, some studies \cite{2018VSE++,chen2020adaptive,wei2020universal,liu2020hal} endeavored to improve the learning objectives. As a seminal work, Faghri \cite{2018VSE++} \textit{et al.} proposed to introduce one on-line hard negative mining (OHNM) strategy into BTR loss, which is very prevailing for ITR. Liu \textit{et al.} \cite{liu2020hal} proposed to tackle hubness problem by imposing heavy punishment on the hard negatives in triplets. Afterwards, Chen \textit{et al.} \cite{chen2020adaptive} further improved the BTR loss by searching for more hard negatives in off-line way to constitute the quintuplet. Overall, the common character of above works is designing constraint strategy for pairwise multi-modal data, whilst our DCL additionally performs diversity estimation especially for each sample. Moreover, we introduce MCL to promote large-scale negative interaction, which leads to more comprehensive diversity estimation in DCL.

%\section{Method}	
\section{Methodology}		
\label{sec:Method}	
\subsection{Overall Framework}
\begin{figure*}[!t]
	\centering
	{
	\includegraphics[height=5.4cm,width=1\linewidth]{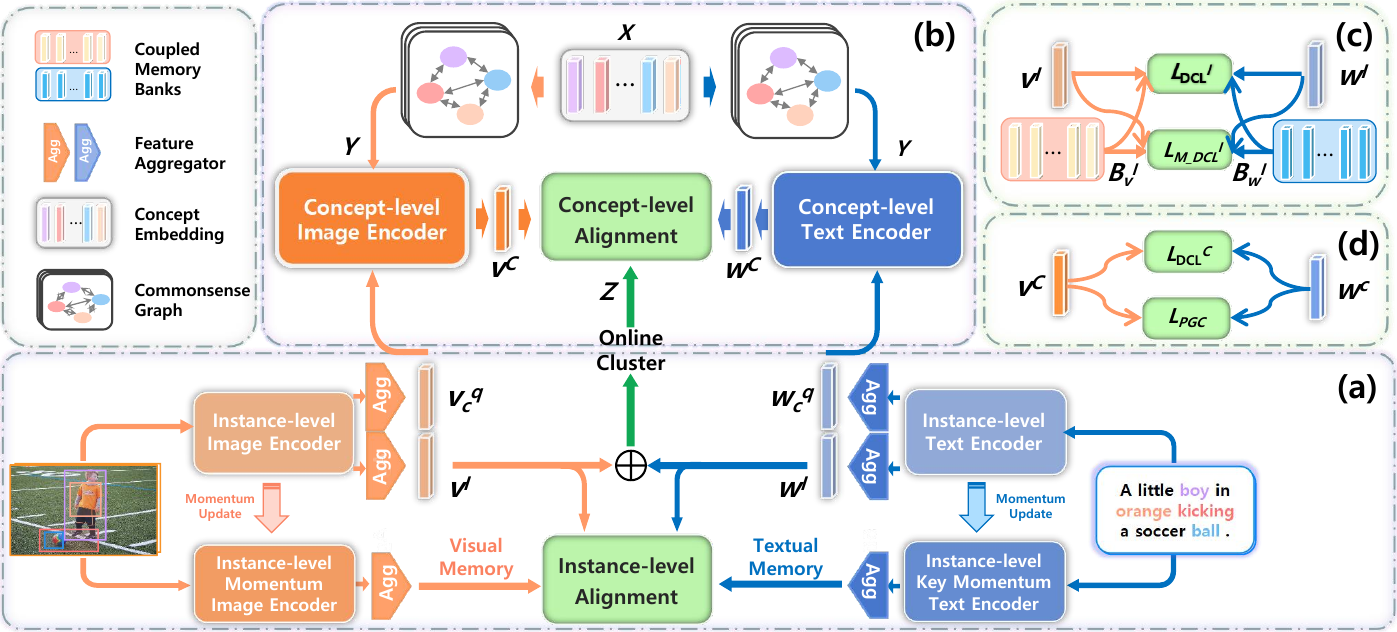}
	}
	\caption{The overall architecture of our proposed CODER model for image-text retrieval. It is composed of an instance-level representation branch (a) and an concept-level one which leverages external knowledge (b). The former branch is optimized by minimizing instance-level DCL loss and memory-based DCL loss (denoted as $L_{DCL}^{I}$ and $L_{M\_DCL}^{I}$, respectively) (c). The other one is learned by employing concept-level DCL loss  $L_{DCL}^{C}$ and online clustering based prototype-guided classification $L_{PGC}$ as objectives (d).}    
	\label{fig.2}
\end{figure*}

The overall framework of our proposed CODER model is illustrated in Figure \ref{fig.2}. In our model, two branches are designed for instance-level and concept-level representation learning. In the instance-level branch (Fig.\ref{fig.2}(a)), image and text features are encoded and aggregated to be $\mathbf{v}^I$ and $\mathbf{w}^I$, momentum encodes are used for the two modalities to serve as coupled memory banks. Instance-level alignment is achieved via employing our proposed diversity-sensitive contrastive loss $L_{DCL}^I$ as well as memory-aided DCL loss $L_{M\_DCL}^I$ (Fig.\ref{fig.2}(c)). As for the concept-level branch (Fig.\ref{fig.2}(b)), statistical commonsense representation (SCC) \cite{Wang2020CVSE} denoted as $\mathbf{Y}$, is adopted as homogeneous feature basis. Query features $\mathbf{v}_C^q$ and $\mathbf{w}_C^q$ are obtained from image and text, respectively. Then concept-level features $\mathbf{v}^C$ and $\mathbf{w}^C$ are obtained by learning to query from feature basis $\mathbf{Y}$. For concept-level alignment (Fig.\ref{fig.2}(d)), except for DCL loss $L_{DCL}^C$, an online-clustering based prototype-guided classification loss $L_{PGC}$ is additionally leveraged.  

\subsection{Instance and Concept Level Representations}
\subsubsection{Instance-level Representation}
\label{sec:Instance-Level_Representation}
For image encoding, we adopt Faster-RCNN \cite{2015FasterRcnn,2018Bottomup} to obtain $L$ region-level features $\{\mathbf{o_l}\}_{l=1}^L$ and then aggregate these features to be a instance-level visual embedding $\mathbf{v}^{I} \in\mathbb{R}^{F}$. Pre-trained BERT \cite{Devlin2019BERTPO} is our textual encoder and $N$ word-level embeddings $\{\mathbf{e_t}\}_{t=1}^T$ are also aggregated to instance-level textual embedding $\mathbf{w}^{I} \in\mathbb{R}^{F}$.
\begin{equation}
	\label{eq:1}
	\begin{aligned}
		\begin{split}
			& \mathbf{v}^{I} = g_{vis}(\left\{\mathbf{o}_{l}\right\}_{n=1}^{L}), \quad \mathbf{w}^{I} = g_{text}(\left\{\mathbf{e}_{t}\right\}_{t=1}^{T}),
		\end{split}	
	\end{aligned}	
\end{equation}
where $g_{vis}(\cdot)$ and $g_{text}(\cdot)$ are visual and textual aggregators.

\subsubsection{Concept-level Representation}
\label{sec:Concept-level_Representation}
The concept-level representations for both modalities are built based on a group of \textit{concepts}. Firstly, we extract $g$ representative concepts from the the texts over the whole image-caption dataset. Afterwards, the GloVE \cite{pennington2014glove} is employed to instantiate these concepts as $\mathbf{X}$. Following \cite{Wang2020CVSE}, graph convolution network (GCN) \cite{kipf2016semi} is utilized to process to produce the statistical commonsense aided concept (SCC) representations $\mathbf{Y}=\left\{ {\mathbf{y}_{1},...,\mathbf{y}_{g}} \right\}$. Please refer to the supplementary materials for more details.

%\textcolor{red}{
To generate concept-level representations, we generate representations ($\mathbf{v}_{C}^q$ and $\mathbf{w}_{C}^q$) by using another group of feature aggregators ($g_{vis}(\cdot)$ and $g_{text}(\cdot)$) to combine local features $\{\mathbf{o_l}\}_{l=1}^L$ and $\{\mathbf{e_t}\}_{t=1}^T$, respectively. Then, as depicted in Figure \ref{fig.2}, $\mathbf{v}_{C}^q$ and $\mathbf{w}_{C}^q$ are fed into concept-level feature encoders, which are taken as input vectors to query from the SCC representations $\mathbf{Y}$. The output scores for different concepts allow us to uniformly utilize the linear combination of the SCC representations to represent both modalities. Mathematically, the concept-level representation $\mathbf{v}^C$ and $\mathbf{w}^C$ can be calculated as:
\begin{equation}		
	\label{eq:vc}
	\begin{aligned}
		\begin{split}	
			& \mathbf{v}^C = \sum_{i = 1}^g {{a}_{i}^v} \mathbf{y}_{i};  \ 		
			{a}_{i}^v = {\frac		
				{e^{\lambda \mathbf{v}_{C}^{I} {\mathbf{W}}^{v} \mathbf{y}_{i}^\mathsf{T}}}
				{\sum_{i = 1}^q e^{\lambda \mathbf{v}_{C}^{I} {\mathbf{W}}^{v} \mathbf{y}_{i}^\mathsf{T}}}}.	\\	
			& \mathbf{w}^C = \sum_{j = 1}^g {{a}_{j}^w} \mathbf{y}_{j};  \ 		
			{a}_{j}^w =\frac		
			{e^{\lambda \mathbf{w}_{C}^{I} {\mathbf{W}}^{w} \mathbf{y}_{j}^\mathsf{T}}}
			{\sum_{j=1}^qe^{\lambda \mathbf{w}_{C}^{I} {\mathbf{W}}^{w}	 \mathbf{y}_{j}^\mathsf{T}}}			
		\end{split}			
	\end{aligned}		
\end{equation}		
\MakeLowercase{where} ${\mathbf{W}}^{v} \in\mathbb{R}^{F \times F}$ and ${\mathbf{W}}^{w} \in\mathbb{R}^{F \times F}$ denote the learnable parameter matrix, $\mathbf{a}_{i}^v$ and $\mathbf{a}_{j}^w$ denote the visual and textual score corresponding to the concept $\mathbf{z}_i$, respectively. $\lambda$ controls the smoothness of the softmax function.
%}

\subsubsection{Coupled Memory Banks Building}
\label{sec:CMCL}
We propose to leverage a couple of dynamic memory banks $B_{v}^{I}$ and $B_{w}^{I}$ to restore more visual and textual embeddings to enlarge the scale of negative samples for both modalities. We follow MoCo \cite{he2020momentum} to obtain instance-level momentum image encoder and text encoder by momentum updating their weights according to the corresponding image and text encoders. Visual or textual instances from the latest training iterations are fed to the momentum encoders to generate visual and textual embeddings, which are restored in coupled memory banks. Such a process can be conveniently implemented via queues.

\subsection{Diversity-Sensitive Contrastive Loss}
\label{sec:DCL}
Estimating the semantic Diversity of instance plays important role in enhancing cross-modal discrimination. Specifically, to describe our diversity-sensitive contrastive loss, we start from diversity estimation, and then introduce our \textit{explicit} diversity-sensitive loss. 

\subsubsection{Diversity Estimation}
For simplicity, we take as example that visual feature $\mathbf{v}_i$ is an anchor sample and $Q$ text features $\mathbf{W} = \{\mathbf{w}_i, \mathbf{w}_2, ..., \mathbf{w}_Q\}$ are to be compared (among which only $\mathbf{w}_i$ is a matching sample for $\mathbf{v}_i$), to illustrate how we estimate diversity of an anchor sample. The cosine similarity of $cosine(\mathbf{v}_i$, $\mathbf{w}_j)$ is defined as $S_{ij}$. We propose a simple but effective metric to estimate the semantic diversity explicitly. 

In joint embedding space, if an anchor sample with low diversity indicates the close similarities between it and numerous negatives, this case is undesired. By contrast, an ideal data distribution space should be more structured and consistent with text-image pair annotations. Intuitively, we propose to quantify the diversity of anchor sample via employing one statistical variable, \textit{i.e.} standard deviation (SD). Concretely, a low-diversity anchor sample has multiple negatives with close distances to it, implying the SD value of cross-modal similarities between it and them will be small. Conversely, the high SD value means an anchor sample has high diversity. Since the SD value between negative cross-modal similarities are proportional to the diversity of anchor, we propose to estimate the semantic diversity explicitly based on SD value. Taking image sample $\mathbf{v}_i$ for instance, the computation process of its diversity value is defined as:	
\begin{equation}	
	\label{eq:hl}	
	\begin{aligned}		
		\begin{split}		
			& SD(\mathbf{v}_i) = \sqrt{E(S^{2}_{ij}) - [E(S_{ij})]^2}, i\neq j; \\& div(\mathbf{v}_i) = 1/ \sigma (\epsilon/SD);  \\
			& div(\mathbf{v}_i) = div(\mathbf{v}_i)/ \max\{ div(\mathbf{v}_1),...,div(\mathbf{v}_Q)\},
			%i\[\ne\]j
			%& SD^{m}(i) = \sqrt{E(S^{m}_{ij}) - [E(S^{m}_{ij})]^2}; \ per^{m}(i) = \sigma (SD^{m})  
			%\\
			%& div(i) = (per^{b}(i) + per^{m}(i)) / 2, 
		\end{split}	
	\end{aligned}	
\end{equation}	
\MakeLowercase{where} $E(\cdot)$ is the mathematical expectation function and $\sigma(\cdot)$ denotes the Sigmoid function that normalizes the reciprocal of SD value to a uniform scale, assuring it vary in a relatively stable range. $div(\mathbf{v}_i)$ denotes the diversity score of $\mathbf{v}_i$ calculated from the candidate textual samples to be compared with. $\epsilon=0.1$ is a tunning parameter. Finally, we divide each diversity score $div_{std}(\mathbf{v}_i)$ by the maximum value of them in mini-batch for normalization. Likewise, the diversity of text sample can be calculated in similar manner. 		

\subsubsection{Diversity-Sensitive Loss} 
As mentioned in Section.\ref{sec:intro}, we aim to highlight the discrepancy among the anchor sample with low-diversity and its negatives. To achieve it, we need to allocate more attention to such cases in order for an optimal alignment model. To begin with, let us term the contrastive objective that insensitive to diversity as $L_{{DCL\_I}}$. Given $\mathbf{V}=\{\mathbf{v}_1,...,\mathbf{v}_N\}$ and $\mathbf{W}=\{\mathbf{w}_1,...,\mathbf{w}_Q\}$, $L_{{DCL\_I}}(\mathbf{V},\mathbf{W})$ can be formulated as: 
\begin{equation}		
	\label{eq:l_DCL_implicit}	
	\begin{aligned}
		%\begin{split}	
		\begin{array}{l}
			{l_{DCL\_I}}(\mathbf{V}, \mathbf{W}) = \frac{\mu}{N}\sum\limits_{n = 1}^N [{log(\sum\limits_{q \ne n} {\exp (\frac{( {S_{nq}}-\gamma)}{\mu}) + 1})} - log({S_{nn}} + 1)];\\
			{l_{DCL\_I}}(\mathbf{W}, \mathbf{V}) = \frac{\mu}{Q}\sum\limits_{q = 1}^Q [{log(\sum\limits_{n\neq q} {\exp (\frac{({S_{qn}}-\gamma)}{\mu}) + 1})} - log({S_{qq}} + 1)];\\
			{L_{DCL\_I}}(\mathbf{W}, \mathbf{V}) = {l_{DCL\_I}}(\mathbf{W}, \mathbf{V}) + {l_{DCL\_I}}(\mathbf{V}, \mathbf{W})
		\end{array}
		%\end{split}		
	\end{aligned}
\end{equation}
\MakeLowercase{where} $\mu$ is a temperature scalar; $\gamma$ is a margin parameter; $N$ is the number of samples within the mini-batch; $S_{nq}= cosine({\mathbf{v}_n}, {\mathbf{w}_q}), S_{qn}= cosine({\mathbf{w}_q}, {\mathbf{v}_n}), S_{nn}= cosine({\mathbf{v}_n}, {\mathbf{w}_n})$ and $S_{qq}= cosine({\mathbf{w}_q}, {\mathbf{v}_q})$ denote the cosine similarities. 

To \textit{explicitly} introduce diversity awareness, we extend the above loss to DCL loss $L_{DCL}$. Mathematically, 	
\begin{equation}
	\begin{aligned}
		\begin{array}{l}
			{L_{DCL}(\mathbf{V},\mathbf{W})} = {l_{DCL}}(\mathbf{W}, \mathbf{V}) + {l_{DCL}}(\mathbf{V}, \mathbf{W}) \\
			{l_{DCL}}(\mathbf{V}, \mathbf{W}) = \frac{\mu}{N}\sum\limits_{n = 1}^N [{log(\sum\limits_{q \ne n} {\exp (\frac{({S_{nq}}-\gamma)}{\mu \cdot div(\mathbf{v}_n)}) + 1})} - log({S_{nn}} + 1)];\\
			{l_{DCL}}(\mathbf{W}, \mathbf{V}) = \frac{\mu}{Q}\sum\limits_{q = 1}^Q [{log(\sum\limits_{n\neq q} {\exp (\frac{({S_{qn}}-\gamma)}{\mu \cdot div(\mathbf{w}_q)}) + 1})} - log({S_{qq}} + 1)];\\
		\end{array}
	\end{aligned}
	\label{loss_p}
\end{equation}
%\begin{equation}	
%\label{eq:l_DCL}	
%\begin{aligned}		
%%\begin{split}		
%L_{DCL} = &\{\frac{1}{N}\frac{1}{\mu }\sum\limits_{i = 1}^N {(log(\sum\limits_{p \ne i} {\exp (\mu (\gamma - {S_{pi}})div(i)) + 1} ) +\\
%& log(\sum\limits_{q \ne i} {\exp (\mu (\gamma - {S_{qi}})div(i)) + 1} ))} - log({S_{ii}} + 1),		\}
%%\end{split}			
%\end{aligned}		
%\end{equation}		
where $div(\mathbf{v}_n)$ and $div(\mathbf{w}_q)$ denotes the diversity of $\mathbf{v}_n$ and $\mathbf{w}_q$, respectively and they are used to adaptively weight the negative samples. 

\subsubsection{DCL Loss Based Cross-Modal Alignment} 
\textbf{Instance-level DCL Loss.} For instance-level representation, two items of DCL loss is employed. First, it is imposed on data pairs in mini-batch, named as $L_{DCL}^{I}$. Secondly, it is imposed on anchor sample in mini-batch and items from coupled memory banks, namely Memory-aided Diversity-sensitive Contrastive Learning (M-DCL) and abbreviated as $L_{M\_DCL}^{I}$. Formally, using $\mathbf{V}^{I}$ and $\mathbf{W}^{I}$ to denote a mini-batch of embeddings $\mathbf{v}^{I}$ and $\mathbf{w}^{I}$, these loss items are defined as: 
\begin{equation}	
	\label{eq:l_DCL_inst}	
	%\begin{aligned}		
	\begin{split}		
		&	 L_{DCL}^{I} = L_{DCL}(\mathbf{V}^{I}, \mathbf{W}^{I}), \\ 
		& 	 L_{M\_DCL}^{I} = L_{DCL}(\mathbf{V}^{I}, B_w^{I}) + L_{DCL}(\mathbf{W}^{I}, B_v^{I}).	
	\end{split}			
	%\end{aligned}		
\end{equation}	
Please note that in Eq.\ref{eq:l_DCL_inst}, because the presence of memory banks, diversity estimation is processed as the average of diversity values at mini-batch level and memory bank level.

\noindent\textbf{Concept-level DCL Loss.} For concept-level representation, we only impose DCL Loss on data pairs in a mini-batch, the concept-level DCL loss is represented as: $L_{DCL}^{C} = L_{DCL}(\mathbf{V}^{C}, \mathbf{W}^{C}) + L_{DCL}(\mathbf{W}^{C}, \mathbf{V}^{C})$.

%	\begin{figure}[t]
%		\begin{center}
%			%\fbox{\rule{0pt}{2in} \rule{0.9\linewidth}{0pt}}
%			\includegraphics[width=0.9\linewidth,height=5.0cm]{Figure3.pdf}
%		\end{center}
%		\setlength{\abovecaptionskip}{+0.1cm}
%		\caption{Conceptual illustration of Prototype-guided Classification (PGC) Loss. Through on-line clustering based on instance-level representations, the output prototype will be assigned to multiple samples as classification label, which captures the structured and general information between semantically similar samples. The samples with the same prototypical class are marked with similar colors.}
%		\label{fig.3}
%		\label{fig:long}
%		\label{fig:onecol}
%	\end{figure}	

\subsection{Prototype-guided Classification Loss}	
\label{sec:PGC_loss}		
In this section, we present a novel Prototype-guided Classification (PGC) loss, which aims to enhance cross-modal discrimination by leveraging the complementary semantics between instance-level and concept-level representations. In particular, we perform K-means \cite{2010Kmeans} clustering in an on-line manner during training based on the summation of instance-level representations $\mathbf{v}^{I}$ and $\mathbf{w}^{I}$, which contains more individual information. We name the output clusters as \textit{prototypes} that are able to capture the shared semantic information between semantically related samples. Accordingly, The prototype ids of image/text instances serve as the pseudo class ids and are taken as supervision $ {\bf{Z}}=\{{{z}_{1}},...,{{z}_{K}}\} $ for concept-level representation learning. Specifically, the PGC loss is formally defined as:		
\begin{equation}	
	\begin{aligned}	
		\begin{split}	
			& {{\mathbf{P}}_{v}}=softmax({{\mathbf{P}}^{C}}{{\mathbf{v}}^{C}}), {{\mathbf{P}}_{w}}=softmax({{\mathbf{P}}^{C}}{{\mathbf{w}}^{C}}), \\
			& L_{PGC} = L_{PGC}^{v} + L_{PGC}^{w}=L_{cls}({\mathbf{P}}_{w}, {\bf{Z}})+L_{cls}({\mathbf{P}}_{v}, {\bf{Z}})
		\end{split}
	\end{aligned}
\end{equation}
%\begin{equation}
%\begin{aligned}
%\begin{split}
%& L_{i}^{m}=-\sum\limits_{n=1}^{k}{1\{i{{d}_{m}}=n\}\log \mathbf{P}_{im}^{n}} ,
%\end{split}
%\end{aligned}
%\end{equation}
%\begin{equation}
%\begin{aligned}
%\begin{split}
%& {{\mathbf{P}}_{w}}=softmax({{\mathbf{W}}^{m}}{{\mathbf{t}}^{C}}), L_{PGC}^{w}=L_{cls}({\mathbf{P}}_{w}, {\bf{Z}})
%\end{split}
%\end{aligned}
%\end{equation}
%\begin{equation}
%\begin{aligned}
%\begin{split}
%& L_{t}^{m}=-\sum\limits_{n=1}^{k}{1\{i{{d}_{m}}=n\}\log \mathbf{P}_{tm}^{n}} ,
%\end{split}
%\end{aligned}
%\end{equation}
where ${{\bf{P}}^{C}}\in {{\mathbb{R}}^{K\times F}}$ is one learnable parameter matrix that outputs distributions over the $K$ category labels for both ${{\bf{v}}^{C}}$ and ${{\bf{w}}^{C}}$. ${{\mathbf{P}}_{v}}\in {{\mathbb{R}}^{K}}$ and ${{\mathbf{P}}_{w}}\in {{\mathbb{R}}^{K}}$ denote probabilities over all labels. $L_{cls}$ denotes the cross-entropy classification loss. 
%Then, we take the linear combination of $L_{PGC}^{v}$ and $L_{PGC}^{w}$, \textit{i.e.} $L_{PGC}=L_{PGC}^{v}+L_{PGC}^{w}$, as the overall PGC loss. 

\subsection{Training and Inference}

%\subsubsection{Training.} 
\textbf{Training Objective.} We deploy the summation of instance-level and concept-level aligning losses as overall training objectives:    
\begin{equation}
	\label{eq:l_all}
	\begin{aligned}
		\begin{split}
			&	L =  \lambda {L}_{DCL}^{I} + {L}_{M\_DCL}^{I} + {L}_{DCL}^{C} + {L}_{PGC},  \\	
			%&	L =  {\lambda }_{1} {L}_{DCL}^{I} + {\lambda }_{2} {L}_{CMCL}^{I} + {\lambda }_{3} {L}_{DCL}^{C} + {\lambda }_{4} {L}_{PGC},  \\
			%& \mathcal{D}_{KL}(\mathbf{a}^t\ \| \ \mathbf{a}^v) = \sum\nolimits_{i = 1}^q {\mathbf{a}_{i}^t} \ log (\frac{\mathbf{a}_{i}^t}{\mathbf{a}_{i}^v}),
		\end{split}			
	\end{aligned}		
\end{equation}		
%where ${\lambda }_{1}=1, {\lambda }_{2}=1, {\lambda }_{3}=1, {\lambda }_{4}=1$ balance the weight of different loss functions. 
%$\mathcal{D}_{KL}$ is the regularization term that align the the visual and textual predicted concept scores.		

%\subsubsection{Inference Scheme.}
\textbf{Inference Scheme.} For inference, we use the weighted summation of instance-level and concept-level cosine similarities to measure the overall cross-modal similarity $S=\beta S(\mathbf{v}^I,\mathbf{w}^I) + (1-\beta) S(\mathbf{v}^C,\mathbf{w}^C)$, where $\beta$ is a balancing parameter.

\begin{table*}[!t]
	\large
	%	\normalsize
	%	\footnotesize
	%	\scriptsize
	%\small	
	%	\small	
	
	%	\setlength{\tabcolsep}{4pt}{
	\centering
	\renewcommand\arraystretch{1.1}		
	%	\begin{center}
	\caption{Comparisons of experimental results on MSCOCO 1K test set and Flickr30K test set. Note that DSRAN \cite{Wen2021LearningDS}, GPO \cite{chen2021learning} and DIME \cite{qu2021dynamic} employ BERT as we use, the rest use inferior text encoders.}			
	\label{tab.1}					
	
	\resizebox{1\textwidth}{!}{		
		%			\addvbuffer[2pt 2pt]{	
		\begin{tabular}{c|c|ccc|ccc|c|ccc|ccc|c}				
			%				\hline \hline		
			\toprule[1.5pt]
			%				\shline	
			\multicolumn{1}{c|}{\multirow{4}{*}{Methods}}  & \multicolumn{1}{c|}{\multirow{4}{*}{Image Encoder}}  & \multicolumn{7}{c|}{MSCOCO}                                                                       & \multicolumn{7}{c}{Flickr30K}                                                                                                                                                   \\	
			%				\hline
			\cline{3-16}
			%				\midrule
			&  & \multicolumn{3}{c|}{Text retrieval}                                       & \multicolumn{3}{c|}{Image Retrieval}                                          & \multicolumn{1}{c|}{\multirow{2}{*}{R@sum}} & \multicolumn{3}{c|}{Text retrieval}                                       & \multicolumn{3}{c}{Image Retrieval}
			& \multicolumn{1}{|c}{\multirow{2}{*}{R@sum}}\\
			
			\multicolumn{1}{c}{}                           & \multicolumn{1}{|c}{}                           & \multicolumn{1}{|c}{R@1} & \multicolumn{1}{c}{R@5} & \multicolumn{1}{c|}{R@10} & \multicolumn{1}{c}{R@1} & \multicolumn{1}{c}{R@5} & \multicolumn{1}{c|}{R@10} & \multicolumn{1}{c|}{}   & \multicolumn{1}{c}{R@1} & \multicolumn{1}{c}{R@5} & \multicolumn{1}{c|}{R@10} & \multicolumn{1}{c}{R@1} & \multicolumn{1}{c}{R@5} & \multicolumn{1}{c|}{R@10} & \multicolumn{1}{c}{}   \\
			%				\hline
			\midrule
			
			DVSA \cite{2015DVSA} (2015) &  R-CNN       & 38.4                    & 69.9                    & 80.5                     & 27.4                    & 60.2                    & 74.8                     & 351.2                   & 22.2                    & 48.2                    & 61.4                     & 15.2                    & 37.7                    & 50.5                     & 235.2                   \\	
			
			m-CNN \cite{2015mcnn} (2015) &  VGG-19    & 42.8   & 73.1  & 84.1  & 32.6  & 68.6   & 82.8   & 384.0    & 33.6   & 64.1    & 74.9   & 26.2    & 56.3 & 69.6  & 324.7   \\	
			
			DSPE \cite{2016Wang} (2016) & VGG-19    &     50.1     & 	79.7      & 89.2                     & 39.6                    & 75.2                    & 86.9              & 420.7                
			& 40.3                    & 68.9                    & 79.9                     & 29.7                & 60.1            & 72.1             &  351.0                   \\
			
			VSE++ \cite{2018VSE++} (2018) & ResNet-152    & 64.7                    & -                       & 95.9                     & 52.0                    & -                       & 92.0                     & -         & 52.9                    & -                       & 87.2                     & 39.6                    & -                       & 79.5                     & -              \\
			
% 			SCO \cite{2018SCO} (2018) & ResNet-152  & 69.9 &	92.9 &	97.5  &	56.7 &	87.5 &	94.8 &	499.3 & 55.5 &	82.0 &	89.3  &	41.1 &	70.5 &	80.1  &	418.5  	\\
			
			SCAN \cite{2018SCAN} (2018) & Faster-RCNN  & 72.7  &	94.8  &	98.4 &	58.8 &	88.4 &	94.8  &	507.9   &	67.4 &	90.3 &	95.8 &	48.6 &	77.7 &	85.2 &	465.0	\\
			
			PVSE \cite{2019PVSE} (2019) & Faster-RCNN   & 69.2                   & 91.6                      & 96.6                     & 55.2                       & 86.5                   & 93.7	& 492.8  & -  & -  & -  & -  & -  & -                \\
			
			VSRN \cite{li2019visual} (2019) & Faster-RCNN 	& 76.2 & 94.8 & 98.2 & 62.8 & 89.7 & 95.1   &	516.8   &	71.3 &	90.6 &	96.0 &	54.7 &	81.8 &	88.2 &	482.6	\\
			
			CVSE \cite{Wang2020CVSE} (2020) & Faster-RCNN 
			& 74.8 &	95.1 &	98.3 &	59.9
			& 89.4 &	95.2  & 512.7  &	73.5 &	92.1 &	95.8  &	 52.9 &	80.4 &	87.8  & 482.5 	\\
			
			IMRAM \cite{Chen2020IMRAMIM} (2020) & Faster-RCNN  	& 76.7	& 95.6	& 98.5	& 61.7	& 89.1	& 95.0 & 516.6
			& 74.1   & 	93.0  & 96.6  &	53.9  & 79.4	& 87.2		& 484.2		\\
			
			%				MMCA \cite{Wei2020MultiModalityCA} & CVPR-20 & 74.8 & 95.6 & 97.7 & 61.6 & 89.8 & 95.2 & 514.7
			%				& 74.2 & 92.8 & 96.4 & 54.8 & 81.4 & 87.8	& 487.4		\\
			
			WCGL \cite{wang2021wasserstein} (2021)	& Faster-RCNN  &	75.4	&	95.5	&	98.6	&	60.8	&	89.3	&	95.3  & 514.9	&	74.8	&	93.3	&	96.8	&	54.8	&	80.6	&	87.5 &	487.8 \\
			
			SHAN \cite{ji2021step} (2021)	& Faster-RCNN &	76.8   	&	96.3	&	98.7	&	62.6	&	89.6	&	95.8 	& 519.5	 &	74.6   	&	93.5	&	96.9	&	55.3	&	81.3	&	88.4  & 490.0 \\	
			
			DSRAN \cite{Wen2021LearningDS} (2021) & Faster-RCNN &	77.1 &	95.3 &	98.1 &	62.9 &	89.9 &	95.3 &	518.6 &	75.3 &	94.4 &	97.6 &	57.3 &	84.8 &	90.9 &	500.3 \\
			
			GPO \cite{chen2021learning} (2021)  & Faster-RCNN  &	78.6 &	96.2 & 98.7 & 62.9 & 90.8  & 96.1 & 523.3 & 78.1 & 94.1 & 97.8 & 57.4 & 84.5 & 90.4 & 502.3 \\
			
% 			GPO \cite{chen2021learning} (2021)  & Faster-RCNN  & 79.7 & 96.4 & 98.9 & 64.8 & 91.4 & 96.3 & 527.5	& 81.7 &	95.4 &	97.6 &	61.4 &	85.9 &	91.5 &	513.5 \\ 
			
			DIME (i-t)\cite{qu2021dynamic} (2021)  & Faster-RCNN &	77.9 &	95.9	&	98.3   & 63.0	&	90.5	&	\textbf{96.2} 	& 521.8	&	77.4 &	95.0	&	97.4   & 60.1	&	85.5	&	91.8	& 507.2 \\
			
			SGRAF \cite{diao2021similarity}  (2021) & Faster-RCNN  &	79.6	&	96.2	&	98.5	&	63.2	&	90.7	&	96.1	&	524.3 &	77.8	&	94.1	&	97.4	&	58.5	&	83.0	&	88.8 &	499.6 \\
			\hline 
			
			CODER  &	Faster-RCNN   &	\textbf{82.1}	 &	\textbf{96.6}	 &	\textbf{98.8}	 &	\textbf{65.5}	 &	\textbf{91.5}	 &	\textbf{96.2}  &	\textbf{530.6}	&	\textbf{83.2}	&	\textbf{96.5}	&	\textbf{98.0}	&	\textbf{63.1}	&	\textbf{87.1}	&	\textbf{93.0} &	\textbf{520.9}	\\			
			%				\hline		\hline		
			\bottomrule[1.5pt]
			%				\shline
	\end{tabular}}	
	%				}	
	%	\end{center}
\end{table*}

\section{Experiments}	
%\subsection{Experimental Setup}

\subsection{Datasets \& Evaluation Metrics}

\textbf{Datasets.} {Flickr30K} \cite{2015Flickr30K} is an image-caption dataset containing 31,783 images, where each image annotated with five sentences. Following \cite{2015mrnn}, we split the dataset into 29,783 training, 1000 validation, and 1000 testing images.The performance evaluation is reported on 1000 testing set. {MSCOCO} \cite{2014COCO} is another image-caption dataset, totally including 123,287 images with each image roughly annotated with five textual descriptions. We follow the public split of \cite{2015DVSA}, including 113,287 training images, 1000 validation images, and 5000 testing images. The result is reported by averaging the results over 5-folds of 1K testing images.

\textbf{Evaluation Metrics.} We utilize two commonly used evaluation metrics,
\textit{i.e.}, R@K and ``R@sum''. Specifically, R@K refers to the percentage of queries in which the ground-truth matchings appear in the top $K$ retrieved results. ``R@sum'' is the summation of all six recall rates of R@K, which provides a more comprehensive evaluation to testify the overall retrieval performance.

\subsection{Implementation Details}
For visual feature encoding, the amount of regions is $L=36$ and the dimension of region embeddings is $2048$. For text encoding, a BERT-base \cite{Devlin2019BERTPO} model is used to extract 768-dimension textual embeddings. The dimension of joint space is set to $F$=1024. For concept-level representation, we adopt 300-dim GloVe \cite{pennington2014glove} trained on the Wikipedia dataset to initialize the semantic concepts. The volume of the concept vocabulary is $g=400$. The size of couple memory banks is set to 4096 and the momentum coefficient is $0.995$. The cluster number $K$ of PGC loss is set to 10000 and 20000 for Flickr30K and MSCOCO dataset, respectively. For the training objective, we empirically set  $\mu=0.1$ and $\gamma=0.3$ in Eq.~\eqref{loss_p}. Our CODER model is trained by Adam optimizer \cite{2014Adam} with mini-batch size of 128. The learning rate is set to be 0.0002 for the first 15 epochs and 0.00002 for the next 15 epochs for both datasets. The balancing parameter in Eq.~\eqref{eq:l_all} is set to $\lambda=3$. For inference, the controlling parameter $\beta$ is equal to 0.9. All our experiments are conducted on a NVIDIA Tesla P40 GPU.

\subsection{Comparisons with state-of-the-art Methods}
The experimental results on two benchmark datasets are listed in Table \ref{tab.1} \footnote{We report our replicated results of \cite{chen2021learning} by using its official code without changing, more discussions are given in the supplementary materials}. As for MSCOCO, we can observe that our CODER is obviously superior to the competitors in most evaluation metrics, which yields a result of 82.1\% and 65.5\% on R@1 for text retrieval and image retrieval, respectively. Specifically, compared with the best competitor SGRAF method, it achieves absolute boost (2.5\%, 0.4\%, 0.3\%) on (R@1, R@5, R@10) for text retrieval. For image retrieval, our method also outperforms other algorithms. Moreover, on Flickr30K dataset, as for the most persuasive criteria, the ``R@sum'' achieved by our model exceeds the second best performance by 13.7\%. These results solidly validate the advance of our method.

\subsection{Ablation Study}	

In this section, we perform a series of ablation studies to explore the impact of the main modules in our CODER method. All the comparative experiments are conducted on the Flickr30K dataset.  

To begin with, we first investigate the effect of each module for instance-level representation. In Table \ref{tab.2}, we employ a framework without adopting coupled memory banks for M-DCL as the baseline (\#1), which utilizes the traditionally prevailing BTR loss \cite{2018VSE++} to perform instance-level alignment instead of our DCL loss. From Table \ref{tab.2}, Comparing \#1 with \#2 based on R@1, the DCL loss brings about 3.2\% improvement for text retrieval and 2.9\% for image retrieval. Moreover, when the coupled memory banks is exploited for M-DCL, Comparing \#3 with \#2, we can obtain additional performance improvement. These results confirm the effectiveness of our proposed DCL learning paradigm for enhancing instance-level discrimination. 

In addition, we explore how the modules for concept-level representation affects the retrieval performance. As shown in Table \ref{tab.2}, comparing \#4 with \#3 based on R@1, the $L_{DCL}^{C}$ loss leads to 0.2\% improvement for text retrieval and 0.2\% for image retrieval. It validates our DCL loss is also effective for concept-level representation learning. Furthermore, when our presented PGC loss is leveraged, comparing \#5 with \#4, it achieves (0.4\%, 0.4\%) boost on (R@1, R@5) for text retrieval and (0.3\%, 0.3\%) boost on (R@1, R@5) for image retrieval. The above results prove our designed concept-level representation learning module can provides more complementary semantics for instance-level one thereby enhancing cross-modal discrimination. 

\setlength{\tabcolsep}{4pt}
\begin{table}
	\huge
	%	\large
	%	\normalsize
	%		\small
	%	\footnotesize
	%	\scriptsize
	%	\tiny
	
	\centering
	\renewcommand\arraystretch{1.1}
	%	\begin{center}
	%		\setlength{\abovecaptionskip}{+0.15cm}
	\caption{Performance comparison of our CODER with different main components on Flickr30K test set. ``Instance-level Alignment'' is abbreviated as ``IA''. ``Concept-level Alignment'' is abbreviated as ``CA''.}
	\label{tab.2}
	
	\resizebox{0.7\textwidth}{!}{
		\begin{tabular}{ccccccccccc}
			%				\hline \hline 
			\toprule[2.5pt]
			%				\multicolumn{1}{c|}{Models}    
			\multicolumn{1}{c}{\multirow{2}{*}{Models}}
			&
			\multicolumn{2}{c}{{IA}}     &                   	                                                                                                                                                   
			%				\multicolumn{1}{c}{PGC loss} 
			\multicolumn{2}{c}{CA} & \multicolumn{3}{c}{Text Retrieval} & \multicolumn{3}{c}{Image Retrieval} \\ 
			%				\hline
			%				\cline{2-9}
			\begin{tabular}[c]{@{}c@{}} \end{tabular} &
			\begin{tabular}[c]{@{}c@{}} \textit{L}$_{M\_DCL}^{I}$ \end{tabular} & \begin{tabular}[c]{@{}c@{}} \textit{L}$_{DCL}^{I}$ \end{tabular} & \begin{tabular}[c]{@{}c@{}} $L_{PGC}$ \end{tabular} &
			\begin{tabular}[c]{@{}c@{}} $L_{DCL}^{C}$ \end{tabular} &
			R@1  & 	R@5   & 	R@10	 	  & R@1  & R@5	  &	R@10      \\ 
			%				\hline
			\midrule
			1 & 	& 	& & 	& 78.7	& 94.5 & 97.0	& 58.6	& 84.8 & 90.1	\\		
			2 &                                                                  & $\checkmark$ 	&                                                               & & 81.9 & 95.6 & 97.9 & 61.5	& 85.8 & 91.8	        \\
			
			3 & $\checkmark$	&   $\checkmark$   &  &             & 82.6             &  96.1       & 98.0        &  62.6          &   86.7   & 92.3              \\
			\midrule	
			4 & $\checkmark$	& $\checkmark$    & 	 &     $\checkmark$       & 82.8             &  96.1     & 98.0        &  62.8        &   86.8    & 92.6           \\
			
			5 & $\checkmark$ & $\checkmark$    &  $\checkmark$  &     $\checkmark$   &   \textbf{83.2}	&	\textbf{96.5}	&	\textbf{98.0}	&	\textbf{63.1}	&	\textbf{87.1}	&	\textbf{93.0}        \\  		
			%				\hline \hline	
			\bottomrule[2.5pt]		
		\end{tabular}			
	}
	%	\vspace{-0.1cm}
\end{table}

	\begin{table}
	\large
	%	\normalsize
	%		
	%	\small
	%	\footnotesize
	%\scriptsize
	%	\tiny
	\centering
	\renewcommand\arraystretch{1.1}
	%	\begin{center}
	%		\setlength{\abovecaptionskip}{+0.15cm}
	\caption{Effect of different configurations of DCL module on Flickr30K test set. Implicit Diversity estimation is abbreviated as ``IE''. Explicit Diversity estimation is abbreviated as ``EE''. ``MB'' means using memory banks for Explicit Diversity estimation.}
	\label{tab.3}	
	\resizebox{0.7\textwidth}{!}{
		\begin{tabular}{ccccccccccc}
			%			\hline \hline 
			\toprule[1.5pt]
			%				\shline
			%				\multicolumn{1}{c|}{Models}     
			\multicolumn{1}{c}{\multirow{2}{*}{Models}}
			&                   	        
			\multirow{2}{*}{$L_{M\_DCL}^{I}$}          &                                           
			%				\multicolumn{1}{c}{PGC loss} 
			\multicolumn{3}{c}{$L_{DCL}^{I}$} & \multicolumn{3}{c}{Text Retrieval} & \multicolumn{3}{c}{Image Retrieval} \\ 
			%				\hline
			%			\cline{2-9}
			\begin{tabular}[c]{@{}c@{}}\end{tabular} &
			\begin{tabular}[c]{@{}c@{}}  \end{tabular} & \begin{tabular}[c]{@{}c@{}} IE \end{tabular} & \begin{tabular}[c]{@{}c@{}} EE \end{tabular} &
			\begin{tabular}[c]{@{}c@{}} MB \end{tabular} &
			R@1  & 	R@5 	& 	R@10     & R@1   & R@5	 & 	R@10       \\
			%			 \hline				
			\midrule
			%				1	& 	& $\checkmark$	& 	& 79.8	& 94.7	& 59.5	& 85.6	\\	\hline	
			%				& 	& 	& 	& 79.8	& 94.7	& 97.2	& 59.5	& 85.6	& 90.8	\\	\hline					
			1 &      &		$\checkmark$                                                               &  &  & 80.3 & 94.8 & 97.3 & 60.2	& 85.3 & 91.1    \\ 
			%				&      &		$\checkmark$                                                               &   & 80.3 & 94.8	& 97.4  & 60.2	& 85.8	& 91.3        \\ 
			2	&     & $\checkmark$ &  $\checkmark$    &  & 81.5             &  95.6       & 97.5       &  61.2          &   85.6  & 91.2                 \\ 
			
			3	&     &  $\checkmark$  &   $\checkmark$   & $\checkmark$   & 81.9            &  95.6     & 97.7         &  61.5       &   85.8       & 91.4            \\
			\hline	
			4	&  $\checkmark$   & $\checkmark$   &            &	 &  82.0             &  95.8     &  97.7             &  61.6          &   86.2       & 92.2             \\ 	
			%				$\checkmark$	&     &  $\checkmark$ &            & 81.0             &  95.2          &   98.0        &  60.3          &   86.1        &    91.4           \\ 
			
			%				& $\checkmark$    &  $\checkmark$  &   $\checkmark$       & 81.9             &  95.6          &   97.6      &  61.5       &   85.8        &    91.3           \\
			
			5	 &  $\checkmark$   & $\checkmark$  &  $\checkmark$  &        & 82.8             &  96.3      &  97.9        &  62.6          &   87.0      &   92.7            \\ 
			%				$\checkmark$ &     & $\checkmark$  &   $\checkmark$         & 82.9              &  \textbf{96.0}          &   98.3       &  62.8          &   87.2        &    92.4           \\ 
			%				\hline		
			6	& $\checkmark$    &  $\checkmark$  &     $\checkmark$   &  $\checkmark$ & \textbf{83.2}	&	\textbf{96.5}	&	\textbf{98.0}	&	\textbf{63.1}	&	\textbf{87.1}	&	\textbf{93.0}         \\  	
			%				$\checkmark$ & $\checkmark$    &  $\checkmark$  &     $\checkmark$   &   \textbf{83.4} &   \textbf{96.1}	&	98.4	&	\textbf{63.5}	&	\textbf{87.3}	&	\textbf{92.9}           \\ 	
			%			\hline \hline		
			\bottomrule[1.5pt]
		\end{tabular}			
	}
	%	\end{center}
	%	\vspace{-0.2cm}
\end{table}

\subsubsection{Impact of Different Configurations of DCL}
In this part, we perform ablation studies to explore the impact of different configurations for the DCL module. 

%To explore the impacts of the size of the coupled memory banks, a group of experiments are carried out. From Table \ref{tab.3}, we can see when the queue size $B_v^{I}$ and $B_w^{I}$ increases from 1024 to 4096, the retrieval performance improves gradually, which demonstrates the large-scale negative interactions for cross-modal learning indeed results in more discriminative cross-modal embedding and performance improvements. Additionally, when the queue size continues to increase, retrieval performance is slightly degraded. We infer the potential reason is that more positive samples are misclassified as negative samples accompanied with the queue size grows.  	

To analyze the impacts of various components in DCL module, we perform a group of experiments and present the results in Table \ref{tab.3}. We take the model adopting $L_{DCL\_I}$ loss in Eq. \ref{eq:l_DCL_implicit} as baseline, named implicit Diversity contrastive loss and abbreviated as ``IE''. As shown in Table \ref{tab.3}, comparing \#2 with \#1 based on R@1, the explicit Diversity estimation additionally leads to 1.2\% improvement for text retrieval and 1.0\% for image retrieval. Moreover, the comparison between \#3 and \#2 validates the introduce of memory bank items in Diversity estimation really matters for alleviating semantic ambiguity. Furthermore, comparing (\#4,\#5,\#6) with (\#1,\#2,\#3), we find the combination of $L_{M\_DCL}$ and $L_{DCL}^{I}$ loss can lead to significant retrieval performance boost, which validates they are mutually beneficial to each other and collaborate to promote discriminative cross-modal embedding learning.

\begin{figure*}[t]
	\centering
	\includegraphics[height=5.3cm,width=0.99\linewidth]{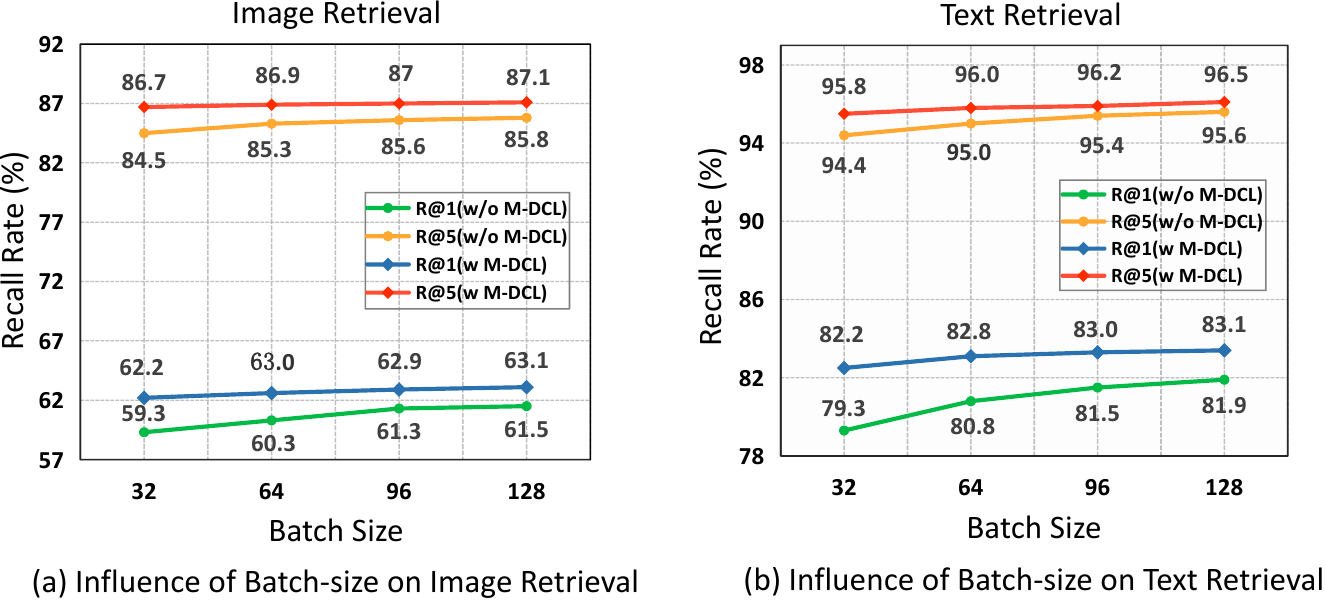}
	\caption{Performance comparison of CODER model with M-DCL and without M-DCL. The model with M-DCL is abbreviated as ``w M-DCL'' and that without M-DCL is abbreviated as ``w/o M-DCL''.}
	\label{fig:fig4}
\end{figure*}

	\textbf{Impact of size in Mini-Batch.} 
	Then, we investigate the impact of size in mini-batch on performance. From Figure \ref{fig:fig4}, we can see that when mini-batch size decreases from 128 to 32, the R@1 metric of the model ``w/o M-DCL'' falls from 61.5\% to 59.3\% for image retrieval,  meanwhile falls from 81.9\% to 79.3\% for text retrieval. By contrast, in the same setting, the R@1 metric of the model ``w M-DCL'' only degrades by 0.9\% and 0.9\% for image retrieval and text retrieval, respectively. These results reveal that, even though the mini-batch size decreases sharply, our CODER with M-DCL can still achieve stable and superior performance, which is achieved by leveraging the coupled memory banks to enlarge interaction with negative samples. Additionally, the insensitivity to mini-batch size indicates our method is able to remain competitive even if the available computation resource is limited.
	
%\begin{figure*}[t]
%		\centering
%		\subfigure[Influence on Text Retrieval]
%		{	
%			\label{fig:fig4(a)}
%			\includegraphics[height=4.85cm,width=0.475\linewidth]{Figure4_a.pdf}}
%		\subfigure[Influence on Image Retrieval]
%		{
%			\label{fig:fig4(b)}
%			\includegraphics[height=4.85cm,width=0.47\linewidth]{Figure4_(b).pdf}}
%		\caption{Performance comparison of CODER model with M-DCL and without M-DCL. The model with M-DCL is abbreviated as ``w M-DCL'' and that without M-DCL is abbreviated as ``w/o M-DCL''.}
%		\label{fig:fig4}
%\end{figure*}

\begin{table}[!t]
		%	\small
		%	\footnotesize
		\scriptsize
		%	\tiny
		
		\centering
		\renewcommand\arraystretch{1.1}
		%	\begin{center}
		%		\setlength{\abovecaptionskip}{+0.15cm}
		\caption{Impact of different clustering number $K$ in PGC loss on Flickr30K test set.}
		\label{tab.4}
		% \centering
		
		\resizebox{0.58\textwidth}{1.38cm}{
			\begin{tabular}{cccccccc}
				% \toprule
				%				\hline \hline	
				\toprule[1.0pt]
				\multirow{2}{*}{$K$} &
				\multicolumn{3}{c}{Text Retrieval} & \multicolumn{3}{c}{Image Retrieval}  \\
				&   R@1 & R@5 & R@10  & R@1 & R@5 & R@10 \\  
				%				\hline
				\midrule
				5000	&82.9	&96.3 &98.0	&\textbf{63.1}	&87.0 &92.7	\\
				%				5000	&83.2	&95.8	&98.3	&63.6	&87.2	&92.7\\
				10000	 &  \textbf{83.2}	&	\textbf{96.5}	&	98.0	&	\textbf{63.1}	&	\textbf{87.1}	&	\textbf{93.0} \\
				%				10000	 &   \textbf{83.4} &   96.1	&	98.4	&	\textbf{63.5}	&	87.3	&	92.9    \\
				15000	&\textbf{83.2}	&96.3  & \textbf{98.2}		&63.0	&\textbf{87.1} & 92.8 \\

				%				15000	&83.2	&96.1	&98.4	&63.3	&\textbf{87.4}	&92.9 \\
				20000	&82.8 & 96.2 &97.9	&62.8	&87.0 & 92.5	\\
				%				20000	&83.0	&\textbf{96.3}	&\textbf{98.5}	&63.0	&87.3	&\textbf{93.1} \\
				%				\hline \hline	
				\bottomrule[1.0pt]	
			\end{tabular}	
		}
		%	\end{center}
		%\label{tab:plain}
		%	\vspace{-0cm}
	\end{table}
	
	\subsubsection{Impact of Different Configurations of PGC Loss}
	
	In this part, we explore the influence of the clustering number $K$ in PGC loss. The corresponding experimental results are listed in Table \ref{tab.4}. It can be seen that the performance is not obviously affected by clustering number, archiving best results at $ K=10000 $. Afterwards, the performance degrades slightly accompanied by the increase of clustering number, which implies the deceasing samples of one prototypical class may weaken the general semantics conveyed by concept-level representations.

%	\begin{figure}[!t]
%		\begin{center}
%			%\fbox{\rule{0pt}{2in} \rule{0.9\linewidth}{0pt}}
%			\includegraphics[width=0.48\linewidth,height=4.8cm]{Figure5.pdf}
%		\end{center}
%		%	\setlength{\abovecaptionskip}{+0.1cm}
%		\caption{Performance comparison of inference speed and recall between different methods. The Kpps on the horizontal axis denote the similarities of how many image-text pairs are calculated per second, the higher the better.}
%		\label{fig.5}
%		\label{fig:long}
%		\label{fig:onecol}
%		%	\vspace{+0.5cm}
%	\end{figure}

	\begin{figure}
		\begin{minipage}[t]{0.5\linewidth}
			\centering
			\includegraphics[height=3.5cm,width=1\linewidth]{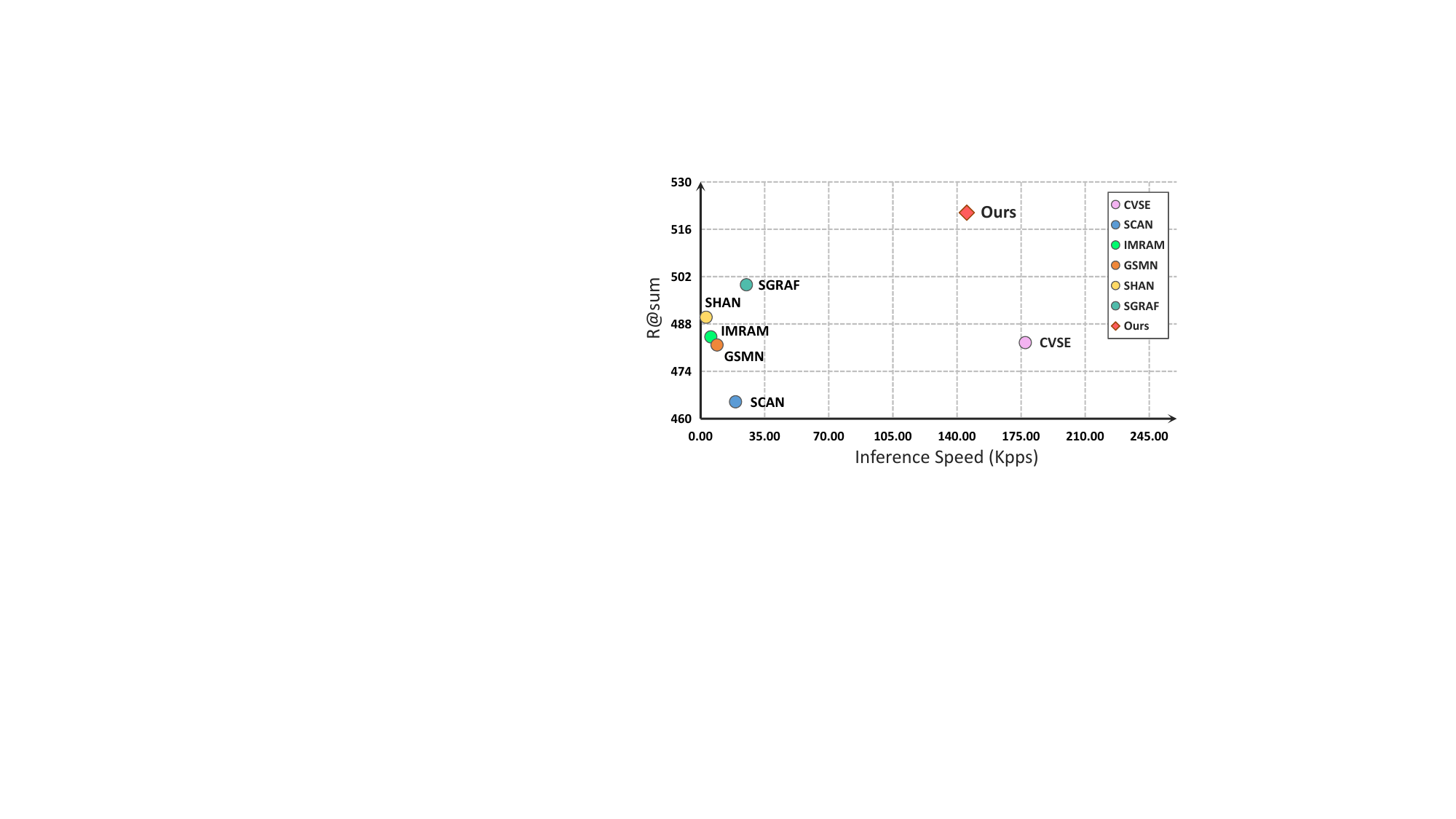}
			\caption{Performance comparison of inference speed and recall between different methods. The Kpps on the horizontal axis denote the similarities of how many image-text pairs are calculated per second, the higher the better.}
			\label{fig.5}
		\end{minipage}%
		\hspace{.1in}
		\begin{minipage}[t]{0.5\linewidth}
			\centering
			\includegraphics[height=3.7cm,width=1\linewidth]{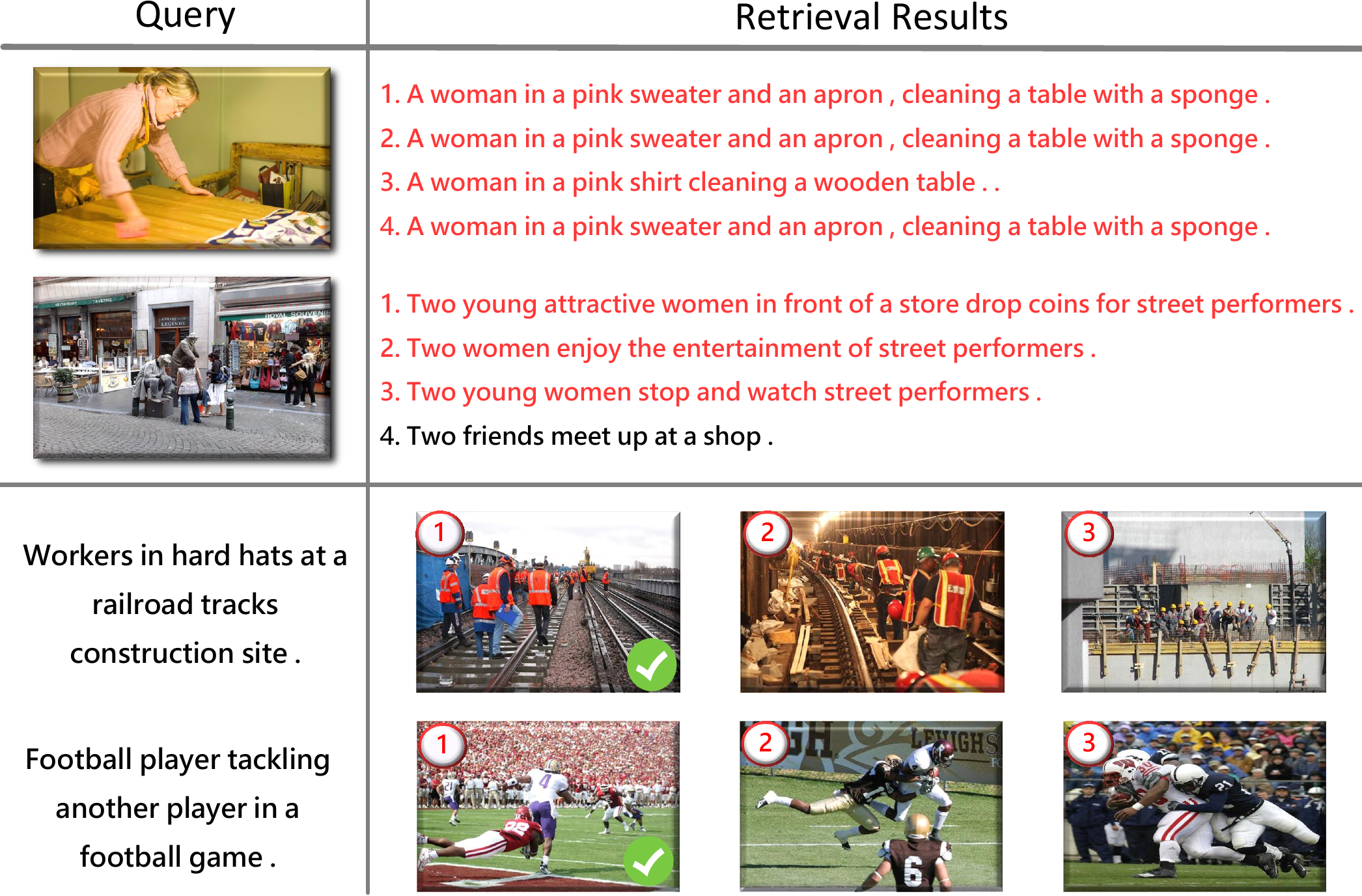}
			\caption{The qualitative bi-directional retrieval results on Flickr30K dataset. For text retrieval, the ground-truth and non ground-truth descriptions are marked in red and black, respectively. For image retrieval, the number in the upper left corner denotes the ranking order, and the ground-truth images are annotated with green check mark.}
			\label{fig.6}
		\end{minipage}
	\end{figure}

	\subsection{Analysis on Accuracy and Efficiency of Model}

	The retrieval latency is also very important in real application scenario, whereas was seldom investigated in previous works. Thus, we report both retrieval recall and consuming time for more comprehensive performance comparisons. To achieve that, we compare our CODER with six leading methods \cite{2018SCAN,Chen2020IMRAMIM,liu2020graph,Wang2020CVSE,ji2021step,diao2021similarity}. Note that the inference time of them are reported by re-implementing their open-sourced codes in the same environment. As shown in Figure \ref{fig.5}, we can see that the inference speed of our method is comparable to CVSE, but its retrieval recall surpasses the latter by a large margin. Besides, in comparison to the best competitor SGRAF \cite{diao2021similarity}, our method surpasses it up to nearly $6\times$ faster for inference, meanwhile achieves considerable advantage over it on ``R@sum'' recall metric. Therefore, our method is superior to these approaches from both perspectives of effectiveness and efficiency.

	\subsection{Retrieval Result Visualization}
	
	To further qualitatively show the performance of our model, in Figure \ref{fig.6}, we select several images and texts as queries to display their retrieval results on Flickr30K dataset. The bidirectional ITR results demonstrate our CODER model can return reasonable retrieval results.

%	\begin{figure}[!t]
%		\begin{center}
%			%\centering
%			%\fbox{\rule{0pt}{2in} \rule{0.9\linewidth}{0pt}}
%			\includegraphics[height=6.cm,width=0.48\linewidth]{Figure6.pdf}
%		\end{center}
%		%		\setlength{\abovecaptionskip}{+0.3cm}
%		\vspace{-0cm}
%		\caption{The qualitative bi-directional retrieval results on Flickr30K dataset. For text retrieval, the ground-truth and non ground-truth descriptions are marked in red and black, respectively. For image retrieval, the number in the upper left corner denotes the ranking order, and the ground-truth images are annotated with green check mark.}
%		\label{fig.6}	
%		\label{fig:long} 
%		\label{fig:onecol}
%		%\end{center}
%		\vspace{-0.1cm}
%	\end{figure}

	\section{Conclusions}
	In this paper, we proposed a Coupled Diversity-Sensitive Momentum Contrastive Learning (CODER) model for image-text retrieval. Specifically, Momentum Contrastive Learning (MCL) is extended to coupled form with dual dynamic modality-specific memory banks to enlarge interactions among instance pairs for cross-modal representation learning. Meanwhile, a novel diversity-sensitive contrastive loss is designed to take semantic ambiguity of sample embedding into account, which flexibly and dynamically allocate attention weights to negative pairs. In parallel, we devise an on-line clustering based strategy to exploit complementary knowledge between hierarchical semantics to promote discriminative feature learning. Furthermore, we systematically studied the impact of multiple components in our model, and its superiority is validated via substantially surpassing state-of-the-art approaches on two benchmarks with very low latency. In the near future, we plan to integrate our proposed learning paradigm into more large-scale vision-language pre-training models.

\clearpage
\bibliographystyle{splncs04}
\bibliography{ref_ECCV2022}

\clearpage

\appendix

\title{-- Supplementary Material -- \\ CODER: Coupled Diversity-Sensitive Momentum Contrastive Learning \\ for Image-Text Retrieval \\ } % Replace with your title

%\begin{comment}
%\titlerunning{ECCV-20 submission ID \ECCVSubNumber} 
%\authorrunning{ECCV-20 submission ID \ECCVSubNumber} 
%\author{Anonymous ECCV submission}
%\institute{Paper ID \ECCVSubNumber}
%\end{comment}

\titlerunning{Supplementary Material of CODER}
\author{}
\institute{}

\maketitle
In this appendix, we provide additional details which were omitted in the main manuscript owing to the limited space. From the perspective of algorithm details, we first give more technical details of representation modules. For instance-level representation, the feature aggregators will be introduced. For concept-level representation, the details of concept selection and the creation of statistical commonsense graph are described. Then, more designed motivation and illustration of our proposed Diversity-sensitive Contrastive Learning (DCL) loss will be given. After that, we also report more experimental results, including the performance of models with different data encoder, influence of different diversity estimation functions, impact of hyper-parameters, data distribution visualization of joint embedding space, performance comparison with different contrastive objectives, and bidirectional image-text retrieval results.

%%%%%%%%% BODY TEXT
\section{Methodology}	
\label{sec:Methodology}

\subsection{Aggregator for Instance-level Representation}
\label{sec:Aggregator}

For simplicity, here we only describe the image feature aggregator for visual modality, since the same goes for the textual branch. Specifically, we employ the Generalized Pooling Operator proposed in \cite{chen2021learning}, which leverages the encoder-decoder architecture to build the image feature aggregator $g_{vis}(\cdot)$: (1) A positional encoding function that turns position index of local features into a vector. (2) A decoding module that takes the positional encoding output to produce pooling weights. 

\textbf{Position Encoder.} To represent each position index $l$ by a dense vector, the positional encoding strategy in Transformer \cite{2017AAN} is adopted:
\begin{equation}
%\label{eq6}
\mathbf{p}_{l}^{i}=\left\{
\begin{aligned}
& {\sin ({u_j},l), if \ i=2j}, \forall i, \\
& {\cos({u_j},l), if \ i=2j + 1},\forall i.
\end{aligned}
\right.
\end{equation}
\MakeLowercase{where} $u_{j} = \frac{1}{{{{10000}^{2j/d_{p}}}}}$ and $d_{p}$ denotes the dimension for positional encoding.

\textbf{Position Decoder.} Given the dense vector ${\mathbf{p}}_{l}\in {{\mathbb{R}}^{d_{p}}}$,, we feed them into a sequence model, which outputs the corresponding pooling weights $\theta = \{\theta\}_{l = 1}^L$. The decoder function contains a bidirectional-GRU (BiGRU) and a two-layer perceptron (MLP):
\begin{equation}
\begin{aligned}
\begin{split}
& \left\{\mathbf{h}\right\}_{l = 1}^L=BiGRU({\left\{\mathbf{p}_{l}\right\}}_{l = 1}^L), \mathbf{\theta}_{k}=MLP(\mathbf{h}_{l})
\end{split}
\end{aligned}
\end{equation}
\MakeLowercase{where} $\mathbf{h}_{l}$ is the hidden states output by Bi-GRU at the position index $k$. Then, we can aggregate the local image features into a holistic instance-level image representation $\mathbf{v}^{I}$:
\begin{equation}
\begin{aligned}
\mathbf{v}^{I}=g_{vis}(\left\{\mathbf{o}_{l}\right\}_{n=1}^{L})=\sum\limits_{l = 1}^L {{\mathbf{\theta}_l} * {\mathbf{o}_l}} 
\end{aligned}
\end{equation}
Similarly, we can obtain the textual feature aggregator $g_{text}(\cdot)$ and global instance-level text representation $\mathbf{w}^{I}$.

\subsection{Concept-level Representation}
\label{sec:Concept-level_Representation}

\subsubsection{Concept Initialization}
Our statistical commonsense knowledge is extracted from certain meaningful concepts and their semantic correlations, which are collected from the texts of the whole image-caption dataset. In order to filter out most meaningless and infrequent concepts, we follow \cite{fang2015captions,2018SCO,Wang2020CVSE} to select the representative words with top-$q$ appearing frequencies in the concept vocabulary, which are roughly categorized into three types, \textit{i.e.}, \textit{Object}, \textit{Motion}, and \textit{Property}. Then, following \cite{Wang2020CVSE}, according to the appearance frequency over dataset, the ratio of the concepts with type of (\textit{Object}, \textit{Motion}, \textit{Property}) is set to (7:2:1). Afterwards, we adopt the glove \cite{pennington2014glove} to initialize them and denote them as $\mathbf{X}$.

\begin{figure*}[t]					
	\begin{center}					
		\includegraphics[width=0.95\linewidth]{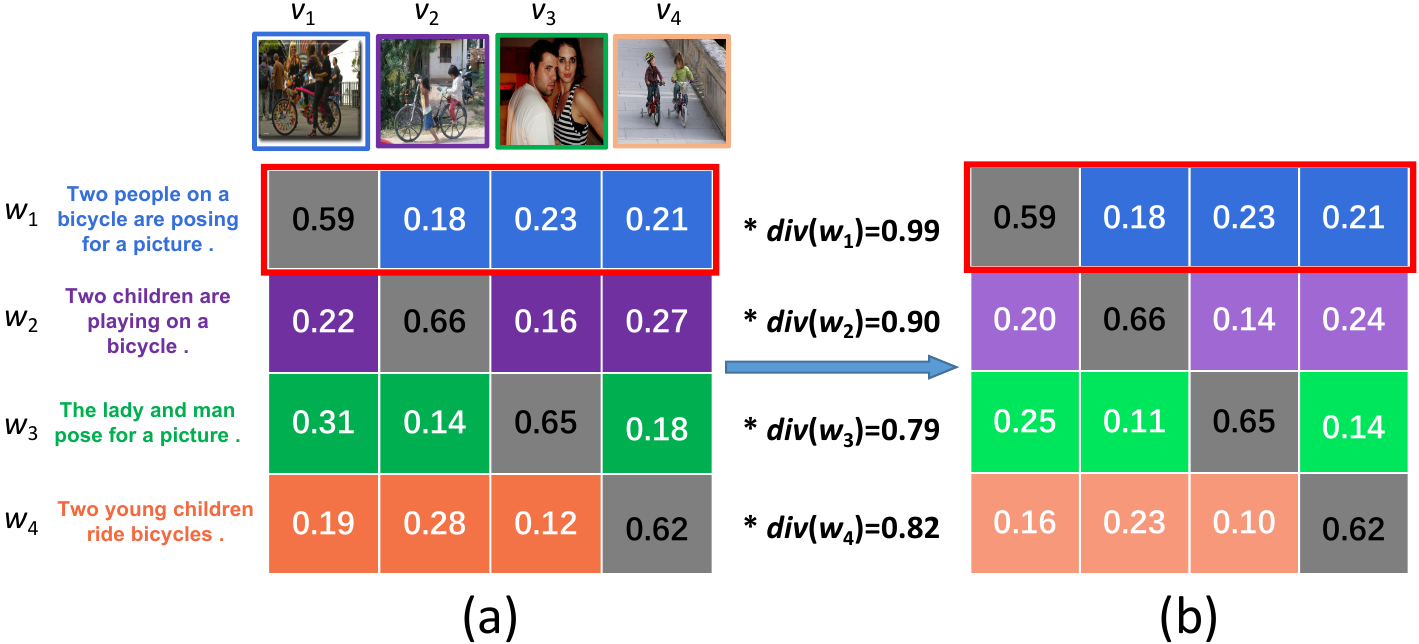}		
	\end{center}				
	\caption{Conceptual illustration of how diversity affects the cross-modal alignment. Sub-figure (a) depicts the image-text similarity matrix and Sub-figure (b) illustrates the similarity matrix being sensitive to the semantic diversity. After the incorporation of diversity score, our DCL loss will focus on handling the samples with high diversity.	
	} 		
	\label{fig.1}				
	\label{fig:long}			
	\label{fig:onecol}			
	\setlength{\abovecaptionskip}{-0.2cm}	
\end{figure*}

\subsubsection{Commonsense Aided Concept Representation.}
To model the statistical commonsense knowledge, we follow \cite{Wang2020CVSE} to utilize the co-occurrence relationship between concepts to build one correlation graph. To be more specific, we construct a conditional probability matrix $\mathbf{P}$ to model the relation between different concepts, with each element $\mathbf{P}_{ij}$ denoting the appearance probability of concept $C_{i}$ when concept $C_{j}$ appears: $ \mathbf{P}_{ij} = \left.{\mathbf{B}_{ij} / {N_{i}}}\right. $, where $\mathbf{B} \in\mathbb{R}^{q \times q}$ is the concept co-occurrence matrix, $\mathbf{B}_{ij}$ represents the co-occurrence times of $C_{i}$ and $C_{j}$, and $N_{i}$ is the appearance times of $C_{i}$ in the corpus. 	

Afterwards, to further prevent the correlation matrix from being over-fitted and improve its generalization ability, we follow \cite{chen2019multi,Wang2020CVSE} to apply binary operation to the rescaled matrix $\mathbf{P}$: 
\begin{equation}
\label{eq:gij}
\mathbf{H}^{sc}_{ij} = \left\{
\begin{aligned}
0 & , & if \ \; \mathbf{P}_{ij} < \ \epsilon, \\
1 & , & if \ \; \mathbf{P}_{ij} \ge \  \epsilon,
\end{aligned}
\right.
\end{equation}
\MakeLowercase{where} $\epsilon$ denotes a threshold parameter filters noisy edges. Given the LCC representations $\mathbf{X}^l$ and statistical commonsense graph $\mathbf{H}^{ss}$, we employ one Graph Convolution Network (GCN) \cite{kipf2016semi} to process them, after one-layer convolution operation, the statistical commonsense aided concept (SCC) representations can be computed as:
\begin{equation}
\label{eq:hl}
\begin{aligned}
\begin{split}
& \mathbf{Y} = \rho (\tilde{\mathbf{A}}_{sc}\mathbf{X}^{l}\mathbf{W}_{sc})
\end{split}
\end{aligned}
\end{equation}	
\MakeLowercase{where} $\tilde{\mathbf{A}}_{sc}=\mathbf{D}_{ss}^{-\frac{1}{2}}\mathbf{H}^{sc}\mathbf{D}_{sc}^{-\frac{1}{2}}+\mathbf{I}$ denotes the normalized symmetric matrix and $\mathbf{W}_{sc}$ is the learnable weight matrix. 

\subsubsection{Commonsense Aided Concept-level Representation.}
To generate concept-level representations, we generate representations ($\mathbf{v}_{C}^q$ and $\mathbf{w}_{C}^q$) by using another group of feature aggregators ($g_{vis}(\cdot)$ and $g_{text}(\cdot)$) to combine local features $\{\mathbf{o_l}\}_{l=1}^L$ and $\{\mathbf{e_t}\}_{t=1}^T$, respectively. Note that the weights of both visual feature aggregators for $\mathbf{v}_{I}$ and $\mathbf{v}_{C}^q$ are shared, and we empirically find this operation helps to make our method converge better. Afterwards, $\mathbf{v}_{C}^q$ and $\mathbf{w}_{C}^q$ are taken as input vectors to query from the SCC representations $\mathbf{Y}$. As consequence, the output scores for different concepts allow us to uniformly utilize the linear combination of the SCC representations to represent both modalities. Mathematically, the concept-level representation $\mathbf{v}^C$ and $\mathbf{w}^C$ can be calculated as:
\begin{equation}		
\label{eq:vc}
\begin{aligned}
\begin{split}	
& \mathbf{v}^C = \sum_{i = 1}^g {{a}_{i}^v} \mathbf{y}_{i};  \ 		
{a}_{i}^v = {\frac		
	{e^{\lambda \mathbf{v}_{C}^{I} {\mathbf{W}}^{v} \mathbf{y}_{i}^\mathsf{T}}}
	{\sum_{i = 1}^q e^{\lambda \mathbf{v}_{C}^{I} {\mathbf{W}}^{v} \mathbf{y}_{i}^\mathsf{T}}}}.	\\	
& \mathbf{w}^C = \sum_{j = 1}^g {{a}_{j}^w} \mathbf{y}_{j};  \ 		
{a}_{j}^w =\frac		
{e^{\lambda \mathbf{w}_{C}^{I} {\mathbf{W}}^{w} \mathbf{y}_{j}^\mathsf{T}}}
{\sum_{j=1}^qe^{\lambda \mathbf{w}_{C}^{I} {\mathbf{W}}^{w}	 \mathbf{y}_{j}^\mathsf{T}}}			
\end{split}			
\end{aligned}		
\end{equation}		
\MakeLowercase{where} ${\mathbf{W}}^{v} \in\mathbb{R}^{F \times F}$ and ${\mathbf{W}}^{w} \in\mathbb{R}^{F \times F}$ denote the learnable parameter matrix, $\mathbf{a}_{i}^v$ and $\mathbf{a}_{j}^w$ denote the visual and textual score corresponding to the concept $\mathbf{z}_i$, respectively. $\lambda$ controls the smoothness of the softmax function.

\subsection{Illustration of Diversity in DCL}
\label{sec:diversity}
In this section, we describe how our proposed semantic \textit{diversity} affects the cross-modal alignment. First, we briefly review the mathematical definitions of diversity and DCL loss defined in the main manuscript. Specifically, we take as example that visual feature $\mathbf{v}_i$ is an anchor sample and $Q$ text features $\mathbf{W} = \{\mathbf{w}_i, \mathbf{w}_2, ..., \mathbf{w}_Q\}$ are to be compared (among which only $\mathbf{w}_i$ is a matching sample for $\mathbf{v}_i$), to illustrate how we estimate diversity of an anchor sample. The cosine similarity of $cosine(\mathbf{v}_i$, $\mathbf{w}_j)$ is defined as $S_{ij}$. Then, the diversity of anchor $\mathbf{v}_i$ ($\mathbf{w}_i$) is defined as:		
\begin{equation}		
\label{eq:diversity}		
\begin{aligned}		
\begin{split}		
& SD(i) = \sqrt{E(S_{ij}) - [E(S_{ij})]^2}, i\neq j, j\in[1,Q]; \\& div_{std}(\mathbf{v}_i)) = 1 / \sigma (\epsilon/SD(\mathbf{v}_i))); \\
& div_{std}(\mathbf{v}_i) = div_{std}(\mathbf{v}_i)/ \max\{ div_{std}(\mathbf{v}_1),...,div_{std}(\mathbf{v}_Q)\},
\end{split}	
\end{aligned}	
\end{equation}
\MakeLowercase{where} $E(\cdot)$ is the mathematical expectation function and $\sigma(\cdot)$ denotes the Sigmoid function that normalizes the reciprocal of SD value to a uniform scale. $div_{std}(\mathbf{v}_i)$ denotes the diversity score of $\mathbf{v}_i$ calculated from the candidate textual samples to be compared with. $\epsilon=0.1$ is a tunning parameter. Lastly, we divide each diversity score $div_{std}(\mathbf{v}_i)$ by the maximum value of them in mini-batch for normalization. Similarly, the diversity of $\mathbf{w}_j$ can be obtained.

Except for the definition above, we also explore another method to define diversity, which is built based on employing \textit{information entropy}. It is commonly used to measure the information volume conveyed by variables. To incorporate cross-modal similarity into information entropy computation, we first utilize softmax function to convert similarity score to probability form. Then, we can use information entropy to estimate the diversity of anchor sample. Formally, the information entropy based diversity of anchor $\mathbf{v}_i$ ($\mathbf{w}_i$) is defined as:	
\begin{equation}		
\label{eq:diversity_ent}		
\begin{aligned}		
\begin{split}		
& P_{ij} = {\frac		
	{e^{{S}_{ij}}} 
	{\sum_{j = 1}^Q e^{{S}_{ij}}}};  \\ &
H(\mathbf{v}_i) = -P_{ij} \cdot \sum_{j = 1}^Q log_{2}({P_{ij}}), i\neq j; \\& div_{ent}(i) = 1 / \sigma (\epsilon/H(i)); \\
& div_{ent}(\mathbf{v}_i) = div_{ent}(\mathbf{v}_i)/ \max\{ div_{ent}(\mathbf{v}_1),...,div_{ent}(\mathbf{v}_Q)\},
\end{split}	
\end{aligned}	
\end{equation}
\MakeLowercase{where} $H(\cdot)$ represents the function for calculating information entropy. $div_{ent}(\mathbf{v}_i)$ denotes the information entropy based  diversity score of $\mathbf{v}_i$ calculated from the candidate textual samples to be compared with. The effect comparison between two types of diversity will be presented in Section \ref{sec:diversity_Estimation}.

Furthermore, given $\mathbf{V}=\{\mathbf{v}_1,...,\mathbf{v}_N\}$ and $\mathbf{W}=\{\mathbf{w}_1,...,\mathbf{w}_Q\}$, based on the diversity score defined in Eq.\ref{eq:diversity}, our proposed DCL loss $L_{DCL}$ is defined as:
\begin{equation}
\begin{aligned}
\begin{array}{l}
{L_{DCL}(\mathbf{V},\mathbf{W})} = {l_{DCL}}(\mathbf{W}, \mathbf{V}) + {l_{DCL}}(\mathbf{V}, \mathbf{W}) \\
{l_{DCL}}(\mathbf{V}, \mathbf{W}) = \frac{\mu}{N}\sum\limits_{n = 1}^N [{log(\sum\limits_{q \ne n} {\exp (\frac{({S_{nq}}-\gamma)}{\mu \cdot div_{std}(\mathbf{v}_n)}) + 1})} - log({S_{nn}} + 1)];\\
{l_{DCL}}(\mathbf{W}, \mathbf{V}) = \frac{\mu}{Q}\sum\limits_{q = 1}^Q [{log(\sum\limits_{n\neq q} {\exp (\frac{({S_{qn}}-\gamma)}{\mu \cdot div_{std}(\mathbf{w}_q)}) + 1})}  - log({S_{qq}} + 1)];\\		
\end{array}
\end{aligned}
\label{loss_p}
\end{equation}
where $div_{std}(\mathbf{v}_n)$ and $div_{std}(\mathbf{w}_q)$ denotes diversity of $\mathbf{v}_n$ and $\mathbf{w}_q$, respectively and they are used to adaptively weight each negative sample. 

Then, we take the similarity measuring matrix of text-to-image as example. As shown in Figure\ref{fig.1}(a), for query sample $\mathbf{w}_1$ with higher diversity, \textit{i.e.} semantic ambiguity, the calculated similarity difference between it and other three mis-matched samples is very small. According to Eq.\ref{eq:diversity}, the diversity of $\mathbf{w}_1$ is larger than those of others. As a result, seeing Figure\ref{fig.1}(b), the similarity ratio between positive pair and negative pairs of $\mathbf{w}_1$ remain unchanged. By contrast, those of other samples all become larger than before. From Eq.\ref{loss_p}, we can know that this adaptive weighting strategy will lead to imposing harder punishment on sample $\mathbf{w}_1$ than others. And it is consistent with our designing principle. 

Note that distinct from previous works \cite{2018VSE++,chen2020adaptive} that focus on mining hard negatives specifically for a single sample, the diversity in DCL is defined from a more holistic view, which is calculated based on the statistical information of data distribution. Consequently, our DCL loss aims at reducing the cross-modal distribution discrepancy, which captures more hierarchical semantic structure in joint space by alleviating the negative influence brought by samples with high diversity. Furthermore, we will display how the diversity in DCL loss affects data distribution in the joint embedding space by t-SNE visualizing in Section \ref{sec:T-SNE}.

\section{Experiments}
\label{sec:Experiments}

% \subsection{More Implementation Details}
% \label{sec:implementation_details}
% All our experiments are implemented on one NVIDIA Tesla P40 GPU with PyTorch toolkit. 

\subsection{More Results and Comparisons for Image-Text Retrieval}
\label{sec:Experiment_results}

The additional experimental results are presented in Table \ref{tab.1}. Note that the results of \cite{chen2021learning} are reported by our replicated number. Since the instance-level representation part of our model is built according to \cite{chen2021learning}, a solid performance of baseline is needed to reasonably evaluate the impact of our contributions. Thus, we report our replicated results by using their open-sourced code with no change, and mark them with $\star$ symbol in Table \ref{tab.1} \& \ref{tab.2}. To further assure the fairness of comparisons, we divide the experiments into two groups. One group of approaches adopt ``Faster-RCNN + BiGRU'' as image and text encoders, meanwhile another group of methods is uniformly built based on ``Faster-RCNN + BERT'' as encoders. The experimental results on Flickr30K test set are presented in Table \ref{tab.1}. First, in contrast to other methods adopting ``Faster-RCNN + BiGRU'' architecture, the ``R@sum'' achieved by our CODER surpasses the second best performance by 13.6\%. Secondly, compared with those employing ``Faster-RCNN + BERT'' for encoding multi-modal data, our method outperforms the best competitor by 13.7\% on the ``R@sum'' metric.

As shown in Table \ref{tab.1}, on MSCOCO 1k test set, our CODER also significantly outperforms all other compared methods. Although employing the 
``Faster-RCNN + BiGRU'' as image and text encoders, there is still a performance gap between CODER and best competitor SMFEA \cite{Ge2021StructuredMF} on the R@sum metric, \textit{e.g.} 4.0\% improvement. Moreover, the retrieval performances on MSCOCO 5K test set are listed in Table \ref{tab.2}. Comparing with best competitor GPO (BERT) \cite{chen2021learning}, our CODER outperforms it by 5.3\% improvement for text retrieval and 1.4\% for image retrieval on R@1 criteria. The above results are obtained under totally fair conditions with the same data encoders, thus they can solidly validate the superiority of our method for image-text retrieval.

\begin{table*}[!t]
	\large
	%	\normalsize
	%	\footnotesize
	%	\scriptsize
	%\small	
	%	\small	
	
	\centering
	\renewcommand\arraystretch{1.18}		
	%	\begin{center}
	\caption{Comparisons of experimental results on MSCOCO 1K test set and Flickr30k test set, employing different image and text encoders (denoted by bold section title).}			
	\label{tab.1}					
	
	\resizebox{1\textwidth}{!}{		
		%			\addvbuffer[2pt 2pt]{	
		\begin{tabular}{c|c|ccc|ccc|c|ccc|ccc|c}				
			%				\hline \hline		
			\toprule[1.5pt]
			%				\shline	
			\multicolumn{1}{c|}{\multirow{4}{*}{Methods}}  & \multicolumn{1}{c|}{\multirow{4}{*}{Image Encoder}}  & \multicolumn{7}{c|}{MSCOCO 1K}                                                                       & \multicolumn{7}{c}{Flickr30K}                                                                                                                                                   \\	
			%				\hline
			\cline{3-16}
			%				\midrule
			&  & \multicolumn{3}{c|}{Text Retrieval}                                       & \multicolumn{3}{c|}{Image Retrieval}                                          & \multicolumn{1}{c|}{\multirow{2}{*}{R@sum}} & \multicolumn{3}{c|}{Text Retrieval}                                       & \multicolumn{3}{c}{Image Retrieval}
			& \multicolumn{1}{|c}{\multirow{2}{*}{R@sum}}\\
			
			\multicolumn{1}{c}{}                           & \multicolumn{1}{|c}{}                           & \multicolumn{1}{|c}{R@1} & \multicolumn{1}{c}{R@5} & \multicolumn{1}{c|}{R@10} & \multicolumn{1}{c}{R@1} & \multicolumn{1}{c}{R@5} & \multicolumn{1}{c|}{R@10} & \multicolumn{1}{c|}{}   & \multicolumn{1}{c}{R@1} & \multicolumn{1}{c}{R@5} & \multicolumn{1}{c|}{R@10} & \multicolumn{1}{c}{R@1} & \multicolumn{1}{c}{R@5} & \multicolumn{1}{c|}{R@10} & \multicolumn{1}{c}{}   \\
			%				\hline
			\midrule
			
			\multicolumn{16}{l}{\textbf{Faster-RCNN + BiGRU}} \\
			%			\midrule
			
			SCAN \cite{2018SCAN} (2018) & Faster-RCNN  & 72.7  &	94.8  &	98.4 &	58.8 &	88.4 &	94.8  &	507.9   &	67.4 &	90.3 &	95.8 &	48.6 &	77.7 &	85.2 &	465.0	\\
			
			VSRN \cite{li2019visual} (2019) & Faster-RCNN 	& 76.2 & 94.8 & 98.2 & 62.8 & 89.7 & 95.1   &	516.8   &	71.3 &	90.6 &	96.0 &	54.7 &	81.8 &	88.2 &	482.6	\\
			
			CVSE \cite{Wang2020CVSE} (2020) & Faster-RCNN 
			& 74.8 &	95.1 &	98.3 &	59.9
			& 89.4 &	95.2  & 512.7  &	73.5 &	92.1 &	95.8  &	 52.9 &	80.4 &	87.8  & 482.5 	\\
			
			%			IMRAM \cite{Chen2020IMRAMIM} (2020) & Faster-RCNN  	& 76.7	& \textbf{95.6}	& 98.5	& 61.7	& 89.1	& 95.0 & 516.6
			%			& 74.1   & 	93.0  & 96.6  &	53.9  & 79.4	& 87.2		& 484.2		\\
			
			MMCA \cite{Wei2020MultiModalityCA} (2020) & Faster-RCNN & 74.8 & \textbf{95.6} & 97.7 & 61.6 & 89.8 & 95.2 & 514.7
			& 74.2 & 92.8 & 96.4 & 54.8 & 81.4 & 87.8	& 487.4		\\
			
			GSMN \cite{liu2020graph} (2020) & Faster-RCNN & 76.1 & \textbf{95.6} & 98.3 & 60.4 & 88.7 & 95.0 & 514.0
			& 74.4 & 91.5 & 95.3 & 54.1 & 79.9 & 86.6	& 481.8		\\
			
			SMFEA \cite{Ge2021StructuredMF} (2021)  & Faster-RCNN &	75.1 &	95.4	&	98.3   & \textbf{62.5}	&	90.1	&	\textbf{96.2} 	& 517.6	&	73.7	&	92.5	&	96.1	&	54.7	&	82.1	&	88.4	& 487.5 \\
			
			WCGL \cite{wang2021wasserstein} (2021)	& Faster-RCNN  &	75.4	&	95.5	&	98.6	&	60.8	&	89.3	&	95.3  & 514.9	&	74.8	&	93.3	&	96.8	&	54.8	&	80.6	&	87.5 &	487.8 \\
			
			GPO (BiGRU) \cite{chen2021learning} (2021) $\star$ & Faster-RCNN  &	76.2 &	95.4 & 98.5 & 60.1 & 89.8  & 95.2 & 515.2 & 74.8 & 93.5 & 97.0 & 55.1 & 83.8 & 89.4 & 493.6 \\
			
			%			SHAN \cite{ji2021step} (2021)	& Faster-RCNN &	76.8   	&	96.3	&	98.7	&	62.6	&	89.6	&	95.8 	& 519.5	 &	74.6   	&	93.5	&	96.9	&	55.3	&	81.3	&	88.4  & 490.0 \\
			
			\midrule
			
			CODER (BiGRU) &	Faster-RCNN   &	\textbf{78.9}	 &	\textbf{95.6}	 &	\textbf{98.6}	 &	\textbf{62.5}	 &	\textbf{90.3}	 &	95.7  &	\textbf{521.6}	&	\textbf{79.4}	&	\textbf{94.9}	&	\textbf{97.7}	&	\textbf{59.0}	&	\textbf{85.2}	&	\textbf{91.0} &	\textbf{507.2}	\\
			
			\midrule	
			\midrule	
			
			\multicolumn{16}{l}{\textbf{Faster-RCNN + BERT}} \\
			%			\midrule
			
			DSRAN \cite{Wen2021LearningDS} (2021) & Faster-RCNN &	77.1 &	95.3 &	98.1 &	62.9 &	89.9 &	95.3 &	518.6 &	75.3 &	94.4 &	97.6 &	57.3 &	84.8 &	90.9 &	500.3 \\
			
			GPO (BERT) \cite{chen2021learning} (2021) $\star$ & Faster-RCNN  &	78.6 &	96.2 & 98.7 & 62.9 & 90.8  & 96.1 & 523.3 & 78.1 & 94.1 & 97.8 & 57.4 & 84.5 & 90.4 & 502.3 \\
			
			DIME (i-t) \cite{qu2021dynamic} (2021)  & Faster-RCNN &	77.9 &	95.9	&	98.3   & 63.0	&	90.5	&	\textbf{96.2} 	& 521.8	&	77.4 &	95.0	&	97.4   & 60.1	&	85.5	&	91.8	& 507.2 \\
			
			%			SGRAF \cite{diao2021similarity}  (2021) & Faster-RCNN  &	79.6	&	96.2	&	98.5	&	63.2	&	90.7	&	96.1	&	524.3 &	77.8	&	94.1	&	97.4	&	58.5	&	83.0	&	88.8 &	499.6 \\
			
			\midrule
			
			CODER (BERT) &	Faster-RCNN   &	\textbf{82.1}	 &	\textbf{96.6}	 &	\textbf{98.8}	 &	\textbf{65.5}	 &	\textbf{91.5}	 &	\textbf{96.2}  &	\textbf{530.6}	&	\textbf{83.2}	&	\textbf{96.5}	&	\textbf{98.0}	&	\textbf{63.1}	&	\textbf{87.1}	&	\textbf{93.0}	&	\textbf{520.9}	\\			
			%				\hline		\hline		
			\bottomrule[1.5pt]
			%				\shline
	\end{tabular}}
\end{table*}

\begin{table*}[!t]
	%	\large
	%			\normalsize
	\footnotesize
	%		\scriptsize
	%	\small	
	%	\small	
	
	\centering
	\renewcommand\arraystretch{1.12}		
	%	\begin{center}
	\caption{Comparisons of experimental results on MSCOCO 5K test set, employing different image and text encoders (denoted by bold section title).}			
	\label{tab.2}					
	
	\resizebox{0.72\textwidth}{2.8cm}{		
		%			\addvbuffer[2pt 2pt]{	
		\begin{tabular}{c|c|ccc|ccc|c}				
			%				\hline \hline		
			\toprule[1.5pt]
			%				\shline	
			\multicolumn{1}{c|}{\multirow{3}{*}{Methods}}  & \multicolumn{1}{c|}{\multirow{3}{*}{Image Encoder}}  & \multicolumn{7}{c}{MSCOCO 5K}                       \\                                                                                                                   
			%				\hline
			\cline{3-9}
			%				\midrule
			&  & \multicolumn{3}{c|}{Text retrieval}                                       & \multicolumn{3}{c|}{Image Retrieval}                                          & \multicolumn{1}{c}{\multirow{2}{*}{R@sum}}                            
			\\
			\multicolumn{1}{c}{}                           & \multicolumn{1}{|c}{}                           & \multicolumn{1}{|c}{R@1} & \multicolumn{1}{c}{R@5} & \multicolumn{1}{c|}{R@10} & \multicolumn{1}{c}{R@1} & \multicolumn{1}{c}{R@5} & \multicolumn{1}{c|}{R@10} & \multicolumn{1}{c}{}     
			\\
			%				\hline
			\midrule
			
			\multicolumn{9}{l}{\textbf{Faster-RCNN + BiGRU}} \\
			%			\midrule
			
			SCAN \cite{2018SCAN} (2018) & Faster-RCNN  & 50.4  &	82.2  &	90.0 &	38.6 &	69.3 &	80.4  &	410.9   \\
			
			VSRN \cite{li2019visual} (2019) & Faster-RCNN 	& 53.0 & 81.1 & 89.4 & 40.5 & 70.6 & 81.1   &	415.7  	\\
			
			%			CVSE \cite{Wang2020CVSE} (2020) & Faster-RCNN 
			%			& 74.8 &	95.1 &	98.3 &	59.9
			%			& 89.4 &	95.2  & 512.7  &	73.5 &	92.1 &	95.8  &	 52.9 &	80.4 &	87.8  & 482.5 	\\
			
			%			IMRAM \cite{Chen2020IMRAMIM} (2020) & Faster-RCNN  	& 76.7	& 95.6	& 98.5	& 61.7	& 89.1	& 95.0 & 516.6		\\
			
			MMCA \cite{Wei2020MultiModalityCA} (2020) & Faster-RCNN & 54.0 & 82.5 & 90.7 & 38.7 & 69.7 & 80.8 & 416.4		\\	
			
			%			GSMN \cite{liu2020graph} (2020) & Faster-RCNN & 76.1 & 95.6 & 98.3 & 60.4 & 88.7 & 95.0 & 514.0 	\\
			
			SMFEA \cite{Ge2021StructuredMF} (2021)  & Faster-RCNN &	54.2 & -	& 89.9	&	\textbf{41.9}  & -	&	\textbf{83.7}	&	-	\\
			
			%			CSCC \cite{zeng2021conceptual}  & Faster-RCNN &	55.6 & 83.6	& 91.2	&	40.8  & 73.2	&	84.3	&	428.7	\\
			
			%			WCGL \cite{wang2021wasserstein} (2021)	& Faster-RCNN  &	75.4	&	95.5	&	98.6	&	60.8	&	89.3	&	95.3  & 514.9	\\
			
			GPO (BiGRU) \cite{chen2021learning} (2021) $\star$ & Faster-RCNN  &	55.2 &	83.1 &	91.0 &	39.3 &	69.9 &	81.1 &	419.6 \\
			
			%			SGRAF \cite{diao2021similarity}  (2021) & Faster-RCNN  &	57.8	&	-	&	91.6	&	41.9	& -	&	81.3	&	- \\
			
			\midrule
			
			CODER (BiGRU) &	Faster-RCNN   &	\textbf{58.5}	 &	\textbf{84.3}	 &	\textbf{91.5}	 &	40.9	 &	\textbf{70.8}	 &	81.4  &	\textbf{427.2}		\\
			
			\midrule	
			\midrule	
			
			\multicolumn{9}{l}{\textbf{Faster-RCNN + BERT}} \\
			
			DSRAN \cite{Wen2021LearningDS} (2021) & Faster-RCNN &	53.7 &	82.1 &	89.9 &	40.3 &	70.9 &	81.3 &	418.2 \\
			
			DIME (i-t) \cite{qu2021dynamic} (2021)  & Faster-RCNN &	56.1 &	83.2 &	91.1 &	40.2 &	70.7 &	81.4 & 422.7	 \\	
			
			GPO (BERT) \cite{chen2021learning} (2021) $\star$ & Faster-RCNN  & 57.3 & 84.5 & 91.6 & 41.1 & 71.9 & 82.6 & 429.0	 \\
			
			%			SGRAF \cite{diao2021similarity}  (2021) & Faster-RCNN  &	79.6	&	96.2	&	98.5	&	63.2	&	90.7	&	96.1	&	524.3 &	77.8	&	94.1	&	97.4	&	58.5	&	83.0	&	88.8 &	499.6 \\
			
			\midrule
			
			CODER (BERT) &	Faster-RCNN   &	\textbf{62.6}	 &	\textbf{86.6}	 &	\textbf{93.1}	 &	\textbf{42.5}	 &	\textbf{73.1}	 &	\textbf{83.3}  &	\textbf{441.3}		\\			
			%				\hline		\hline		
			\bottomrule[1.5pt]
			%				\shline
	\end{tabular}}
\end{table*}

\subsection{Impact of Different Functions for Diversity Estimation}
\label{sec:diversity_Estimation}
In this section, we explore the effect of different diversity estimation functions. As shown in Table \ref{tab.3}, the experimental results based on estimation functions of $div_{std}(\cdot)$ and $div_{ent}(\cdot)$ are listed. For comparison, the results without diversity estimating are also presented. From Table \ref{tab.3}, we can see our proposed two types of diversity estimation functions can both bring about substantial performance boost. It further validates our train of thought for diversity estimation is reasonable. Besides, the performance of model using $div_{ent}(\cdot)$ is slightly inferior to that with $div_{std}(\cdot)$. The potential reason may be that the softmax function adopted in $div_{ent}(\cdot)$ will made the original data distribution of cross-modal similarity to be more smooth.

\begin{table}[!t]
	%	\small
	%	\footnotesize
	\scriptsize
	%	\tiny
	\setlength\tabcolsep{3pt}
	\centering
	\renewcommand\arraystretch{1.1}
	%	\begin{center}
	%		\setlength{\abovecaptionskip}{+0.15cm}
	\caption{Impact of different diversity estimation functions in DCL loss on Flickr30K test set. Explicit diversity estimation is abbreviated as ``EE''.}			
	\label{tab.3}		
	% \centering		
	\resizebox{0.8\textwidth}{1.4cm}{		
		%		\resizebox{0.4\textwidth}{1.45cm}{
		\begin{tabular}{ccccccccc}	
			\toprule[1.0pt]	
			\multirow{2}{*}{EE}  &
			{Diversity Estimation} &		
			\multicolumn{3}{c}{Text Retrieval} & \multicolumn{3}{c}{Image Retrieval}  \\
			&  Function  &  R@1 & R@5 & R@10  & R@1 & R@5 & R@10 \\  
			%				\hline
			\midrule
			$\times$  &- & 81.5 & 95.8 &98.1	&61.3	&85.7 & 91.0  \\
			
			\midrule	
			$\checkmark$ &  $div_{std}(\cdot)$	 &   \textbf{83.2}	&	\textbf{96.5}	&	\textbf{98.0}	&	\textbf{63.1}	&	\textbf{87.1}	&	\textbf{93.0}	\\
			
			$\checkmark$ & $div_{ent}(\cdot)$	&83.0 & 96.2 &\textbf{98.1}	&62.4	&86.9 & 92.5	\\
			%				20000	&83.0	&\textbf{96.3}	&\textbf{98.5}	&63.0	&87.3	&\textbf{93.1} \\
			%				\hline \hline	
			\bottomrule[1.0pt]	
		\end{tabular}	
	}
\end{table}

\begin{figure}[!t]
	\begin{center}
		%\centering
		\includegraphics[width=1\linewidth]{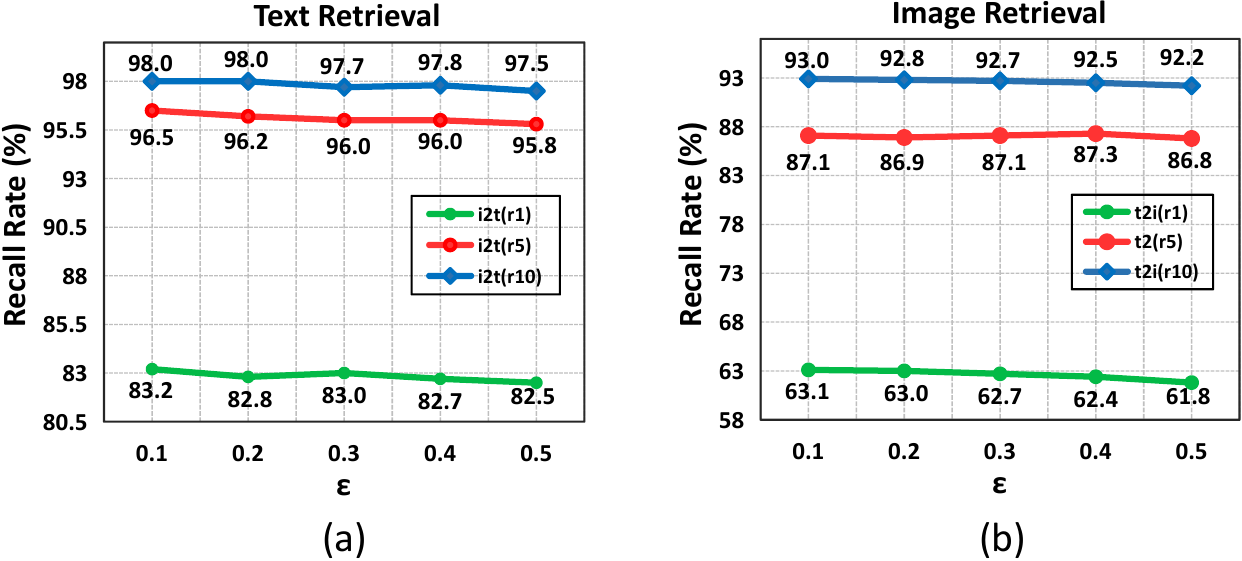}
	\end{center}
	\caption{Impact of varied controlling parameters $\epsilon$ on
		Flickr30K test set. Sub-figure (a) shows image-to-text retrieval performance with different values of $\epsilon$ in DCL loss. Sub-figure (b) depicts the corresponding text-to-image retrieval performance.	
	}
	%	\vspace{-0.3cm}
	\label{fig:fig8}
\end{figure}

\subsection{Hyper-Parameter Analysis for Diversity Estimation}
In this part, we investigate the affect of controlling parameter $\epsilon$ of diversity in Eq.\ref{eq:diversity} on retrieval performance. As shown in Figure \ref{fig:fig8}, with the variant $\epsilon$, the retrieval results vary moderately, indicating our model is robust to $\epsilon$ within a proper range. Additionally, the increase of $\epsilon$ value implicates the narrower variation range of diversity score. Thus, from Figure \ref{fig:fig8}, we can infer the proper sensitiveness of DCL loss on parameter $\epsilon$ also leads to performance gain. Overall, these results reveal that diversity plays critical role in DCL loss for learning more discriminative cross-modal embeddings.

\begin{figure}[t]
	\centering
	\subfigure[]
	{
		\includegraphics[width=0.465\columnwidth]{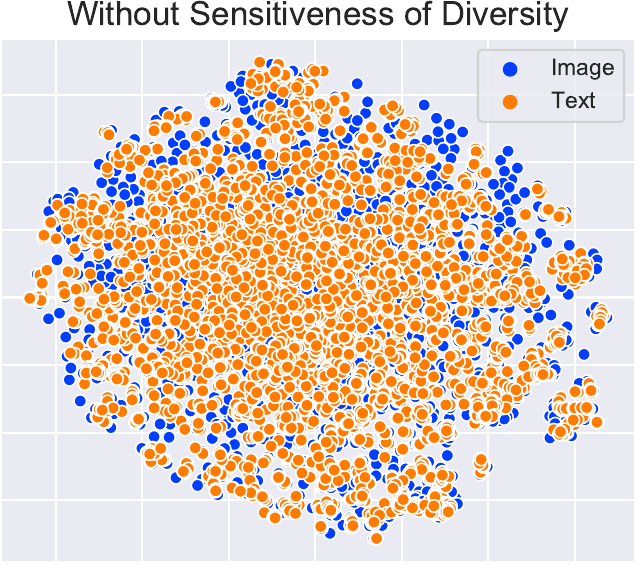}
		\label{fig:fig9(a)}
	}
	% 	\hspace{0.1in}
	\subfigure[]
	{
		\includegraphics[height=5.05cm,width=0.46\columnwidth]{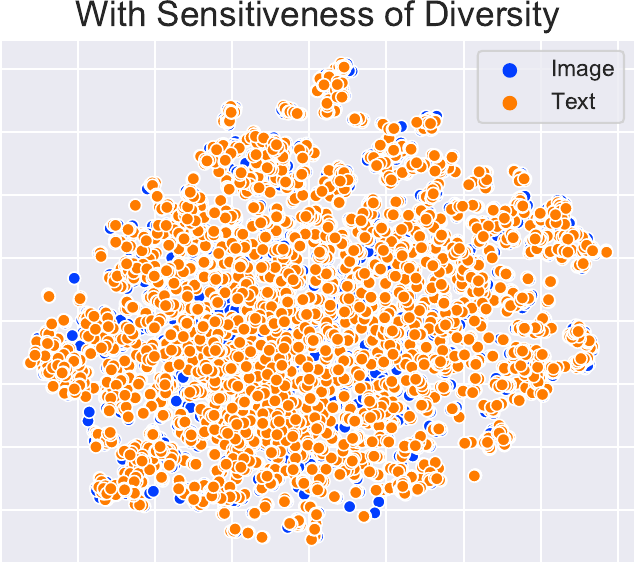}
		\label{fig:fig9(b)}
	}
	\caption{T-SNE visualization of the image-text representations generated by (a) baseline model with $L_{DCL\_I}$ loss and (b) full CODER model on Flickr30K test set (1000 images and 5000 texts).}
	\label{fig:fig9}
	%	\vspace{-0.7cm}
\end{figure}	

%\begin{figure}[t]
%	\centering
%	\begin{subfigure}[t]{\columnwidth}
%		\centering
%		\includegraphics[height=4.9cm,width=0.8\columnwidth]{Figure_8_a.pdf}
%		\caption{}
%		\label{fig:fig8(a)}
%		\vspace{0.5cm}
%	\end{subfigure}
%	
%	\begin{subfigure}[t]{\columnwidth}
%		\centering
%		\includegraphics[height=4.7cm,width=0.8\columnwidth]{Figure_8_b.pdf}
%		\caption{}
%		\label{fig:fig8(b)}
%	\end{subfigure}
%	\caption{Impact of varied controlling parameters $\epsilon$ on
%		Flickr30K test set. Sub-figure (a) shows image-to-text retrieval performance with different values of $\epsilon$ in DCL loss. Sub-figure (b) depicts the corresponding text-to-image retrieval performance.	
%		}
%	\vspace{-0.3cm}
%	\label{fig:fig8}
%\end{figure}

\begin{figure*}[!t]
	\begin{center}
		%\centering
		\includegraphics[width=1\linewidth]{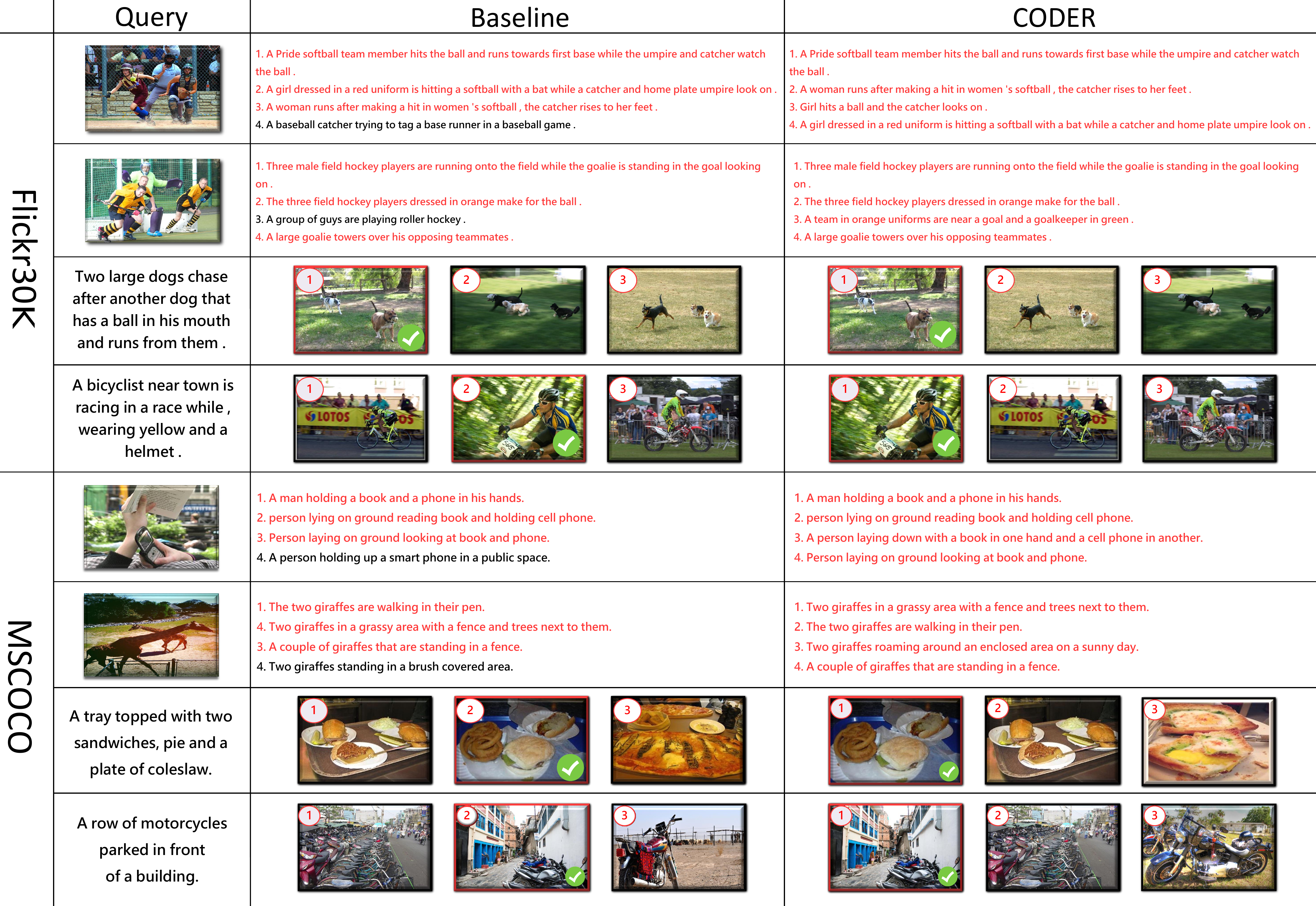}
	\end{center}
	%		\setlength{\abovecaptionskip}{+0.3cm}
	%	\vspace{-0cm}
	\caption{The qualitative bi-directional retrieval results on Flickr30k and MSCOCO datasets. For text retrieval, the ground-truth and non ground-truth describing sentences are marked in red and black, respectively. For image retrieval, the number in the upper left corner denotes the ranking order, and the ground-truth images are annotated with green check mark.}
	\label{fig.11}	
	\label{fig:long} 
	\label{fig:onecol}
	%\end{center}
	\vspace{-0.4cm}
\end{figure*}

\subsection{T-SNE Visualization of Cross-Modal Representation}
\label{sec:T-SNE}

To better understand how our DCL loss affects the cross-modal joint embedding space, we utilize t-SNE \cite{maaten2008visualizing} to visualize the learned representations from Flickr30K test set, including 1000 images and 5000 texts. Specifically, Figure \ref{fig:fig9}(a) displays the feature distribution of the baseline model (referring to the model \#1 in Table 4 defined in the main manuscript, employing the $L_{DCL\_I}$ loss as learning objective), and those of full CODER model is illustrated in Figure  \ref{fig:fig9}(b). We can see that the data distribution in Figure  \ref{fig:fig9}(b) is obviously more desirable than that in Figure  \ref{fig:fig9}(a), which lies in two main points: 1) The distribution discrepancy between images and texts is alleviated significantly. 2) The learned joint space is characterized by being structured and hierarchical rather than being irregular and scattered. Considering the unique difference between both models is the varied configuration of DCL loss, we believe the main factor improving the data distribution is the combination between coupled memory banks ($L_{M\_DCL}^{I}$) and diversity estimation. Benefiting from the large-scale negative interactions from the former, we can achieve more accurate diversity estimation for DCL. It is able to regularize the joint embedding space by alleviating the influence of sample with high diversity, such as some visual instances existing on the left side of Figure  \ref{fig:fig9}(a).

\subsection{Retrieval Result Visualization}
\label{sec:retrieval_result}

To further validate the effectiveness of our method, in Figure \ref{fig.11}, we choose several images and texts as queries and exhibit their retrieval results. Note that we take CODER adopting BTR loss \cite{2018VSE++} instead of our DCL loss as baseline model. As shown in Figure \ref{fig.11}, comparing with baseline, the CODER model with the aid of DCL loss is able to return better image-text retrieval results.

%{\small
%	\bibliographystyle{splncs04}
%	\bibliography{ref_ECCV2022}
%}

\end{document}